\documentclass{article}

    \PassOptionsToPackage{numbers, sort&compress}{natbib}
 \usepackage[preprint]{neurips_2026}

\usepackage[utf8]{inputenc} %
\usepackage[T1]{fontenc}    %
\usepackage[colorlinks=true,allcolors=black]{hyperref}       %
\usepackage{url}            %
\usepackage{booktabs}       %
\usepackage{amsfonts}       %
\usepackage{nicefrac}       %
\usepackage{microtype}      %
\usepackage{xcolor}         %

\usepackage{tikz,pgfplots,pgf}
\usetikzlibrary{arrows.meta,positioning,calc}

\usepackage{subcaption}
\usepackage{caption}
\usepackage{graphicx}
\usepackage{amsmath}
\usepackage[table]{xcolor}
\usepackage{tabularx}
\usepackage{multirow}
\usepackage{mathrsfs}
\usepackage{makecell}
\usepackage{enumitem}

\usepackage{wrapfig}
\usepackage{xspace}
\usepackage{amsmath,amsfonts,bm}

\def\eqref#1{equation~\ref{#1}}

\def\1{\bm{1}}

\def\vc{{\bm{c}}}

\def\vf{{\bm{f}}}

\def\vr{{\bm{r}}}
\def\vs{{\bm{s}}}

\def\vx{{\bm{x}}}

\def\mJ{{\bm{J}}}

\def\mR{{\bm{R}}}
\def\mS{{\bm{S}}}

\def\mW{{\bm{W}}}

\DeclareMathAlphabet{\mathsfit}{\encodingdefault}{\sfdefault}{m}{sl}
\SetMathAlphabet{\mathsfit}{bold}{\encodingdefault}{\sfdefault}{bx}{n}

\newlength{\figurewidth}
\newlength{\figureheight}

\newlength{\spysize}

\usetikzlibrary{spy,calc,datavisualization.formats.functions, patterns}
\usetikzlibrary{positioning}
\newcommand{\ours}{Smol-GS\xspace}

\newcommand{\MLP}{\mathrm{MLP}}
\newcommand{\mipnerf}{{\sc Mip-NeRF~360}\xspace}
\newcommand{\tandt}{{\sc Tanks and Temples}\xspace}
\newcommand{\deepblending}{{\sc Deep~Blending}\xspace}

\newcommand{\ie}{\textit{i.e.}\xspace}
\newcommand{\eg}{\textit{e.g.}\xspace}

\newcommand{\figpath}{./jpeg/fig}

\renewcommand{\paragraph}[1]{\noindent\textbf{#1}~~}

\usepackage[capitalize,nameinlink]{cleveref}

\crefname{section}{Sec.}{Secs.}
\crefname{appendix}{App.}{Apps.}
\crefname{algorithm}{Alg.}{Algs.}
\crefname{figure}{Fig.}{Figs.}

\title{Smol-GS: Compact Representations for \\ Abstract 3D Gaussian Splatting}

\author{%
  Haishan Wang \quad Mohammad Hassan Vali \quad Arno Solin \\
  ELLIS Institute Finland and Aalto University, Espoo, Finland \\
  \texttt{\{haishan.wang, mohammad.vali, arno.solin\}@aalto.fi}
}

\begin{document}
\maketitle
\begin{abstract}
We present \ours, a novel method for learning compact representations for 3D Gaussian Splatting (3DGS). Our approach learns highly efficient splat-wise features to model 3D space, which capture abstracted cues, including color, opacity, transformation, and material properties. We propose octree-derived positional encoding, which explicitly models spatial locality and enhances representation efficiency. We further apply entropy-based compression to exploit feature redundancy and compress splat coordinates using a recursive voxel hierarchy. This design enables orders-of-magnitude reductions in storage while preserving representation flexibility. \ours achieves state-of-the-art compression performance on standard benchmarks with high-level rendering quality.
\sloppy
\end{abstract}
\noindent
Project page: {\small\url{https://aaltoml.github.io/Smol-GS}}

\begin{center}
    \centering\footnotesize
    \captionsetup{type=figure}
    \vspace*{-.5em}
    \resizebox{\textwidth}{!}{
    \begin{minipage}{1.3\textwidth}    
    \begin{tikzpicture}[inner sep=0]

      \tikzstyle{arrow} = [
        -{Stealth[length=3mm,width=2mm]},
        shorten >= 8pt,
        shorten <= 8pt,        
        line width=0.9pt,
        draw=black!70,
        rounded corners=6pt,
        black!30
      ]      

      \node[inner sep=0] (a) at (0,0) {
        \includegraphics[width=.25\textwidth]{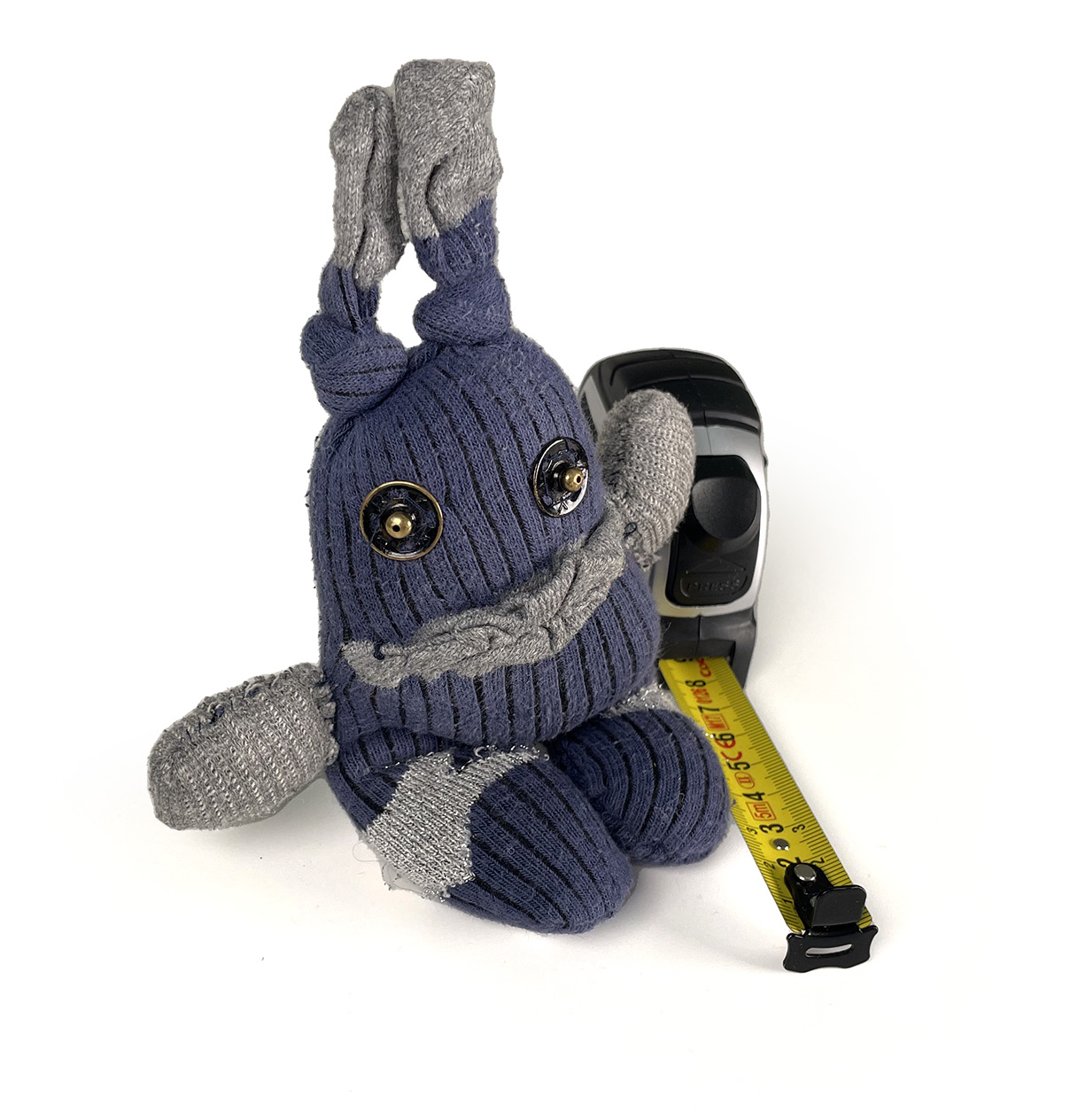}};
        
      \node[inner sep=0] (b) at (.23\textwidth,0) {
        \includegraphics[width=.25\textwidth]{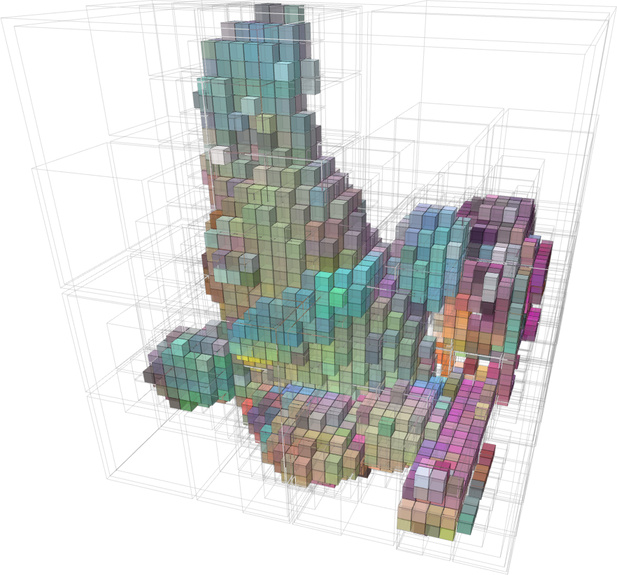}};

      \node[inner sep=0] (c) at (.50\textwidth,0) {
        \includegraphics[width=.22\textwidth]{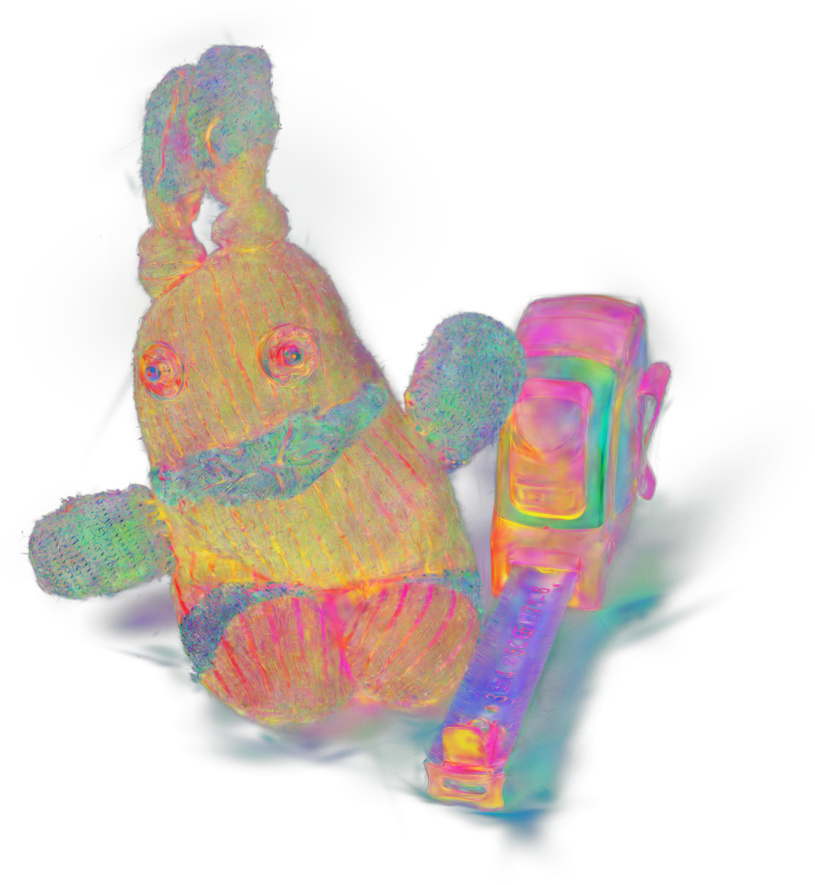}};

      \node[inner sep=0] (d) at (.75\textwidth,0) {
        \includegraphics[width=.22\textwidth]{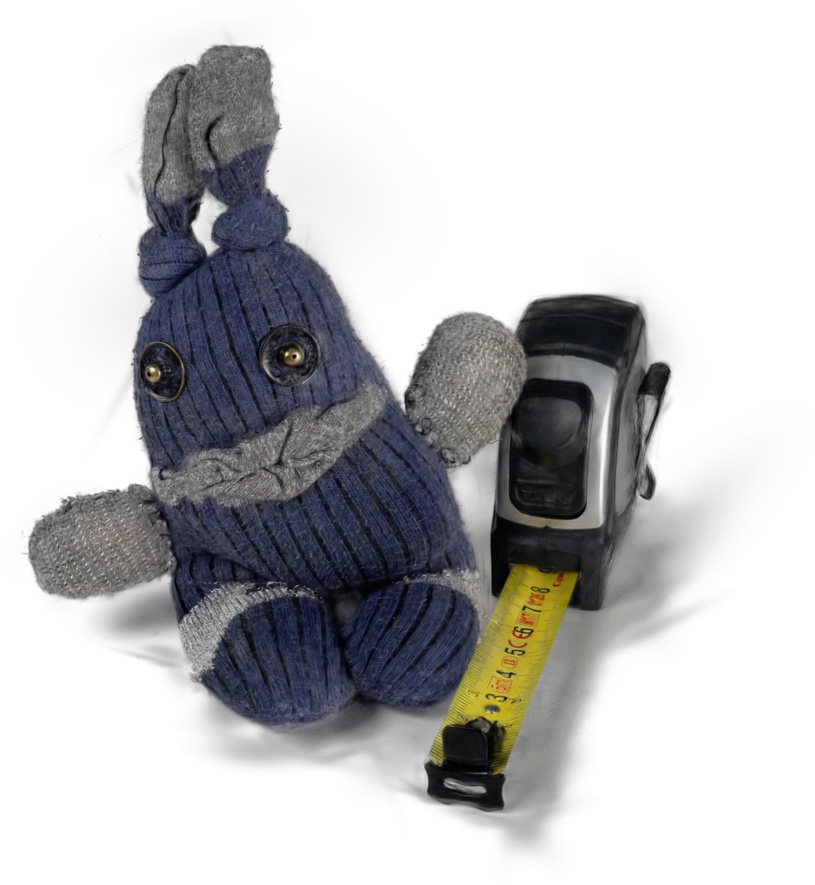}};

      \node[anchor=north west,align=left] (L0) at ($(a) + (-1.5cm,-1.8cm)$) 
        {Training view};
      \node[anchor=north west,align=left] (L1) at ($(b) + (-1.8cm,-1.8cm)$) 
        {Coordinate compression \\ and positional encoding};
      \node[anchor=north west,align=left] (L2) at ($(c) + (-1.5cm,-1.8cm)$) 
        {Splat features\\ \mbox{abstracted} away};
      \node[anchor=north west,align=left] (L3) at ($(d) + (-1.5cm,-1.8cm)$) 
        {Rendered\\ \mbox{reconstruction}};

      \draw[arrow] ($(L0.north east)+(0,-3pt)$) -- ($(L1.north west)+(0,-3pt)$);
      \draw[arrow] ($(L1.north east)+(0,-3pt)$) -- ($(L2.north west)+(0,-3pt)$);
      \draw[arrow] ($(L2.north east)+(0,-3pt)$) -- ($(L3.north west)+(0,-3pt)$);

    \end{tikzpicture}
    \end{minipage}}

    \captionof{figure}{\textbf{\ours} learns a compact representation of the 3D scene that {\em (i)}~stores coordinates in an efficient octree structure, {\em (ii)}~learns local positional encodings, and {\em (iii)}~abstracts away view-dependent splat features such as color, shape, and material cues. On \mipnerf, this gives state-of-the-art results in size/PSNR: 4.87~MB/27.61~dB. 
    }
    \label{fig:teaser}
    \vspace*{-1em}
\end{center}%
\begin{figure*}[t!]
    \centering\tiny
    \setlength{\figureheight}{.095\textwidth}        
    \setlength{\figurewidth}{1.5\figureheight}

    \setlength{\spysize}{0.48\figureheight}
    \foreach \method/\methodname [count=\i from 0] in {gt/{\strut GT},3dgs/{\strut 3DGS-30K},smolgs-base/{\strut \ours}} {        

    \begin{tikzpicture}[
        inner sep=0,
        image/.style = {inner sep=0pt, outer sep=1pt, minimum width=\figurewidth, anchor=north west, text width=\figurewidth}, 
        node distance = 1pt and 1pt, every node/.style={font= {\tiny}}, 
        label/.style = {font={\footnotesize\bf\vphantom{p}},anchor=south,inner sep=0pt}, 
        spy using outlines={rectangle, magnification=2, size=\spysize}
        ] 

    \foreach \scene/\scenename [count=\j from 0] in {bicycle/Bicycle,bonsai/Bonsai,counter/Counter,garden/Garden,room/Room} {
    \ifnum\j=0
        \node [image] (img-\i\j) at (0,-\i*\figureheight) 
          {\includegraphics[height=\figureheight]{\figpath/qualitative_comparison/\method_\scene}};
      \else
        \pgfmathtruncatemacro{\prev}{\j-1}
        \node [image] (img-\i\j) at ($(img-\i\prev.north east)+(\spysize,0)$)
          {\includegraphics[height=\figureheight]{\figpath/qualitative_comparison/\method_\scene}};
      \fi

      \ifnum\i=0
        \node[label,anchor=south,font=\strut\bf] at (img-0\j.north) {\itshape \scenename};
      \fi

      \ifnum\j=0
        \node[label,rotate=90,anchor=south,font=\strut] at (img-\i0.west) {{\bf \methodname}};
      \fi

      \coordinate (spypoint-\j-a) at ($(img-\i\j.north east) + (0.00\figurewidth,-0.02\figurewidth)$);
      \coordinate (spypoint-\j-b) at ($(img-\i\j.south east) + (0.00\figurewidth,+0.02\figurewidth)$);      
      
    }

    \coordinate (spot-1-a) 
      at ($(img-\i0.south west) + (0.45\figurewidth,.45\figurewidth)$); %
    \spy[blue,magnification=2] on 
      (spot-1-a) in node[anchor=north west] at (spypoint-0-a);
    \coordinate (spot-1-b) 
      at ($(img-\i0.south west) + (0.7\figurewidth,.3\figurewidth)$); %
    \spy[red,magnification=4] on 
      (spot-1-b) in node[anchor=south west] at (spypoint-0-b);

    \coordinate (spot-2-a) 
      at ($(img-\i1.south west) + (0.17\figurewidth,.64\figurewidth)$); %
    \spy[blue,magnification=4] on 
      (spot-2-a) in node[anchor=north west] at (spypoint-1-a);
    \coordinate (spot-2-b) 
      at ($(img-\i1.south west) + (0.55\figurewidth,.58\figurewidth)$); %
    \spy[red,magnification=1.5] on 
      (spot-2-b) in node[anchor=south west] at (spypoint-1-b);  

    \coordinate (spot-3-a) 
      at ($(img-\i2.south west) + (0.08\figurewidth,.55\figurewidth)$); %
    \spy[blue,magnification=4] on 
      (spot-3-a) in node[anchor=north west] at (spypoint-2-a);
    \coordinate (spot-3-b) 
      at ($(img-\i2.south west) + (0.3\figurewidth,.22\figurewidth)$); %
    \spy[red,magnification=1.5] on 
      (spot-3-b) in node[anchor=south west] at (spypoint-2-b);   

    \coordinate (spot-4-a) 
      at ($(img-\i3.south west) + (0.3\figurewidth,.48\figurewidth)$); %
    \spy[blue,magnification=4] on 
      (spot-4-a) in node[anchor=north west] at (spypoint-3-a);
    \coordinate (spot-4-b) 
      at ($(img-\i3.south west) + (0.5\figurewidth,.50\figurewidth)$); %
    \spy[red,magnification=2] on 
      (spot-4-b) in node[anchor=south west] at (spypoint-3-b);  

    \coordinate (spot-5-a) 
      at ($(img-\i4.south west) + (0.35\figurewidth,.23\figurewidth)$); %
    \spy[blue,magnification=4] on 
      (spot-5-a) in node[anchor=north west] at (spypoint-4-a);
    \coordinate (spot-5-b) 
      at ($(img-\i4.south west) + (0.5\figurewidth,.4\figurewidth)$); %
    \spy[red,magnification=2] on 
      (spot-5-b) in node[anchor=south west] at (spypoint-4-b);  
    
    \end{tikzpicture}
    \vspace{0.1em}
    }
    \caption{\textbf{Qualitative results on the {\bfseries\scshape Mip-NeRF 360} data set.} \ours provides high-fidelity reconstructions competitive with vanilla 3DGS-30K. \ours preserves sharp shadow edges  and material effects (specular highlights, glass, flat walls) while using an orders-of-magnitude smaller model. We include quantitative summary results in \cref{table:quantitative_benchmark}.
    }
    \label{fig:qualitative}
    \vspace{-1.5em}
\end{figure*}

\section{Introduction}
\label{sec:intro}

3D Gaussian Splatting (3DGS, \cite{kerbl20233dgs}) is a paradigm for novel view synthesis that models a scene as a set of rasterized Gaussian splats. It achieves real-time speed and high rendering quality through rasterization techniques \cite{lassner2021pulsar}. However, 3DGS suffers from memory inefficiency; it often requires several gigabytes of storage with millions of splats to represent complex scenes. This issue of large memory footprint has been tackled extensively in recent works \cite{chen2024hac,chen2025hac++,liu2024hemgs, navaneet2024compgs,lee2024compact3d,niedermayr2024compressed3d,papantonakis2024reducing3d,morgenstern2024sog,fan2024lightgaussian,wang2024contextgs,mallick2024taming3dgs,fang2024minisplatting,zhang2025gaussianspa,pateux2025bogauss,ren2024octree,girish2024eagles,wang2025nsvqgs,ye2025gsplat}. These methods improve compactness to varying degrees, yet the coordinate data and high-dimensional appearance cues remain hard to store efficiently at scale.

Two main directions dominate current 3DGS compression research: {\em (i)}~Signal-processing approaches quantize splat attributes and deliver considerable storage savings \citep{navaneet2024compgs,lee2024compact3d,fan2024lightgaussian,papantonakis2024reducing3d,niedermayr2024compressed3d,wang2025nsvqgs}; {\em (ii)}~Learning-based approaches compress by sharing structure across splats, with anchor-offset hierarchies from Scaffold-GS \citep{lu2024scaffoldgs} forming the backbone of the most competitive results \citep{chen2024hac,chen2025hac++,liu2024hemgs,wang2024contextgs}. However, signal-processing pipelines do not fully exploit local spatial correlations and repeated patterns among splats, and anchor-offset designs can introduce transparent or invisible offsets and extra heuristics. These approaches often sidestep coordinate compression because reconstruction is sensitive to spatial errors. We revisit this premise in our work (see \cref{fig:teaser}) that hinges on the realization that keeping the spatial structure explicit (but efficiently packed) and abstracting the splat-wise visual features away can result in highly efficient compression. 

In this paper, we propose \ours by disentangling geometry from appearance and introducing positional encodings to splats. Our method learns a compact representation with splat-wise abstract features and locality-aware embeddings. Inspired by prior works \citep{lu2024scaffoldgs,schnabel2006octree,chen2024hac}, in \ours coordinates are stored with an occupancy-octree encoding and entropy coding; features (color, opacity, transformation, and material cues) and scaling controllers are quantized and entropy-coded with distributions predicted from hashed spatial descriptors. Our design avoids anchor-offset complications, which introduce storage redundancy and rendering latency. Qualitative (\cref{fig:qualitative}) and quantitative (\cref{fig:quantitative} and \cref{table:quantitative_benchmark}) results and inference speed comparison (\cref{table:fps_comparison}) show that \ours achieves higher reconstruction quality and rendering speed while pushing model size down to the smallest among strong baselines.

\paragraph{Contributions} Our contributions can be summarized as:
\begin{itemize}[leftmargin=1.5em]
\item We introduce \ours, a 3DGS compression model that learns an explicit and highly efficient representation. Our design avoids the structural overhead of anchor-based methods, exploits the representational redundancy to compress scenes further, and accelerates rendering.
\item We propose a novel positional encoding $\vf_{oct}$ using an octree hierarchy. These locality-aware features enhance reconstruction quality.
\item Experiments on standard benchmarks show that \ours substantially reduces the storage requirements for 3DGS while maintaining high reconstruction fidelity and rendering speed, competitive with state-of-the-art approaches, making \ours well-suited for real-time applications.
\end{itemize}

\section{Related work}
\label{sec:related}
\noindent\textbf{3D Gaussian splatting~~}
The 3DGS paradigm represents a scene as a collection of Gaussian splats initialized via structure-from-motion (SfM, \cite{schonberger2016sfm}), based on computer graphics research \cite{westover1991splatting,zwicker2002ewa}. Each splat is associated with a location, covariance (via scale and rotation), opacity, color, and view-dependent material properties (spherical harmonics, SH). The explicit primitive and Gaussian rasterization allow real-time rendering, while the continuous representation yields high-quality reconstructed images. These advantages make 3DGS a promising framework for broad applications, such as SLAM \cite{yan2024gsslam,keetha2024splatam,huang2024photoslam,hu2024cgslam}, 3D editing \cite{chen2024gaussianeditor,wu2024gaussctrl,shao2024control4d,zhuang2024tipeditor}, and VR \cite{tang2023dreamgaussian,zielonka2025drivable,jiang2024vr,zhang2024physdreamer}.

\paragraph{Compaction vs.\ attribute compression}
Efforts to address the challenging memory footprint of storing the splat-wise parameters has mainly focused on two lines of research. \emph{Compaction} reduces the number of splats via pruning and regularization, often guided by opacity, scale, importance estimation \cite{mallick2024taming3dgs,fang2024minisplatting,zhang2025gaussianspa,pateux2025bogauss,ren2024octree}. This can save memory but risks losing thin structures and requires heuristics. \emph{Attribute compression} keeps the splat set but encodes attributes more efficiently, balancing rate and fidelity. For example,  
vector-quantization \cite{van2017neural_vq,gersho2012vector,vali2023stochastic} based methods utilize clustering algorithms \cite{lloyd1982least,linde1980algorithm} to compress per-splat attributes (\eg, color/SH, opacity, scale/rotation) using codebooks \cite{navaneet2024compgs,lee2024compact3d,fan2024lightgaussian,papantonakis2024reducing3d,niedermayr2024compressed3d}.  
However, traditional tools cannot eliminate the information redundancy hidden in continuity and correlation in 3D space.\looseness-1

\paragraph{Learned compact representations}
Anchor-offset hierarchies introduced by Scaffold-GS~\cite{lu2024scaffoldgs} condense local neighborhoods into shared anchor embeddings and reconstruct splat features with small MLPs. Follow-ups add stronger coding: HAC++~\cite{chen2025hac++} and HEMGS~\cite{liu2024hemgs} use entropy models, and ContextGS~\cite{wang2024contextgs} leverages autoregressive anchors to reduce storage further. These methods are effective because they share structure across splats and model statistics. However, the anchors always introduce a certain ratio of transparent offsets, and the recovery of splat coordinates requires the composition of three factors: anchor location, relative coordinates, and scaling control. These design drawbacks, stemming from heuristic complications, limit the compression ratio and flexibility for downstream applications. Additionally, the methods typically avoid coordinate compression due to quality sensitivity, which limits the efficiency of compression.

\paragraph{Coordinate compression and geometry coding}
Point-cloud geometry compression shows that hierarchical occupancy enables compact, often lossless storage of 3D locations via octrees \citep{schnabel2006octree,que2021voxelcontext,fu2022octattention,cui2023octformer}. This perspective suggests that 3DGS coordinates---despite their importance for fidelity---can also be stored efficiently using an occupancy-octree with entropy coding~\cite{huffman2007method}, leaving appearance and view-dependent effects to be learned by low-dimensional features.\looseness-1

\ours is different from previous anchor-offset designs by {\em (i)}~adopting an individual neural splat formulation that eliminates anchor-offset bookkeeping; {\em (ii)}~encoding splat coordinates with an occupancy-octree combined with entropy coding; and {\em (iii)}~introducing an octree-derived positional encoding to enhance splat representation efficiency. \ours aims to enhance the data exploitation capability of learning-based methods.
\section{Methods}
\label{sec:methods}
\begin{figure*}[t!]
  \centering\footnotesize
  \resizebox{\textwidth}{!}{
  \begin{minipage}{1.6\textwidth}
  \begin{tikzpicture}[>={Stealth[round]}]

    \newcommand{\stack}[3]{ %
      \foreach \i in {0,1,2}
        \node[rounded corners=1pt,draw=black!60,inner sep=2pt,fill=white] 
          (#2-\i) at ($(#1)+(\i*3pt,-\i*3pt)$)
            {\includegraphics[width=1cm]{#3}};
    }

    \tikzstyle{arrow} = [
      -{Stealth[length=3mm,width=2mm]},
      shorten >= 4pt,
      line width=0.9pt,
      draw=black!70,
      rounded corners=6pt
    ]

    \tikzstyle{block} = [
      draw=black!60,
      fill=gray!10,
      rounded corners=3pt,
      inner sep=5pt,
      text width=7em,
      align=center,
      minimum height=3em
    ]

\tikzset{
  curlybrace/.style={
    decorate,
    decoration={brace, amplitude=4pt},
    thick,
    draw=black!70
  },
  curlytext/.style={
    align=left,
    anchor=west,
    text width=5em
  }
}

\newcommand{\curlywithtext}[4]{%
  \draw[curlybrace] (#1) -- node[curlytext, name=#3, midway, xshift=1pt]{#4} (#2);
}
\coordinate (brace-bottom) at (.58\textwidth, 2);
\coordinate (brace-top) at (.58\textwidth, 3);

\curlywithtext{brace-bottom}{brace-top}{curlybrace}{
  $\bm{\Sigma} \in \mathbb{R}^{3{\times}3}$\\
  $\vc \in \mathbb{R}^3$\\
  $o \in \mathbb{R}$
}

    \node[minimum size=6em] (splats) at (0,0) {};
    \node at (splats) {
      \includegraphics[width=.2\textwidth]{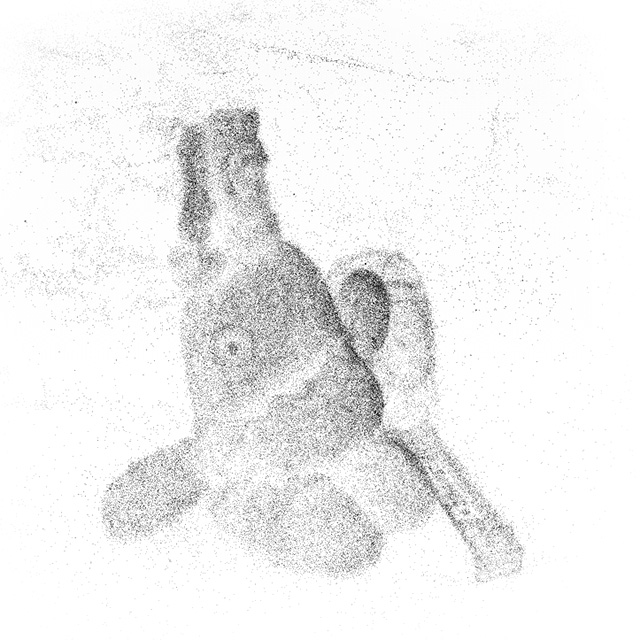}};

    \node[block] (quantization) at (.2\textwidth,0) {Quantization};

    \node[minimum size=6em] (octree) at (.39\textwidth,0) {};    
    \node at (octree) {
      \includegraphics[width=.2\textwidth]{\figpath/teaser/poko-oct}};

    \node[block] (rasterization) at (.6\textwidth,0) {Rasterization ($\alpha$-blending)};

    \stack{.8\textwidth,0}{render}{\figpath/pipeline/poko-render}

    \stack{.8\textwidth,-1.65cm}{gt}{\figpath/pipeline/poko-gt}

    \node[block] (adc) at (.3\textwidth,-.13\textwidth) {Adaptive Density Control};

    \node[block] (sfm) at (0,.15\textwidth) {SfM \\ point~cloud};

    \node[block, text width=6.5em] (features) at (.46\textwidth,2.5) {Feature generation} ;
    \node[above=1pt of features,align=left,yshift=0pt]{${\color{blue}\MLP_o}\;,\;{\color{blue}\MLP_c}\;,\;{\color{blue}\MLP_r}$\\\quad${\color{blue}\MLP_s}\;,\;{\color{blue}\MLP_{oct}}$};
    \draw[arrow] (sfm) -- node[right,yshift=4pt]{Initialize} (splats);
    \draw[arrow] (splats) --  (quantization);
    \draw[arrow,shorten >=18pt] (quantization) -- (octree);
    \draw[arrow,shorten <=20pt] (octree) -- node[right,yshift=6pt]{$\hat{\vx}$} (rasterization);
    \draw[arrow] (rasterization) -- (render-0);
    \draw[arrow] (rasterization) |- (adc);
    \draw[arrow,shorten >=20pt] (adc) -| (splats);
    \draw[arrow,red] (quantization) -- ++(0,-1.2cm) node[red,xshift=-7pt,yshift=-3pt]{$\lambda_q \left[ \mathcal{L}_\text{NLL}(\vf) + \mathcal{L}_\text{NLL}(\vs) \right]$};
    \draw[arrow,shorten <=47pt] ($(octree)+(-20pt,0pt)$) |- node[left,align=left,yshift=-10pt]{Quantized\\attributes\\$\hat{\vx} \in \mathbb{R}^3$\\$\hat{\vf} \in \mathbb{R}^8$\\$\hat{\vs} \in \mathbb{R}^3$}(features);
\draw[arrow,shorten >=9pt] (features) -- (curlybrace);
\draw[arrow,red] ($(curlybrace)+(5pt,-10pt)$) -- ++(1.0cm,0) node[right,red,xshift=-2pt]{$\lambda_o \mathcal{L}_o$};
\draw[arrow,shorten <=5pt] ($(curlybrace)+(-14pt,-12pt)$) -- (rasterization);
    
    \node[left= 2pt of splats,align=left,yshift=-3pt]{Splat\\attributes\\${\color{blue}\vx} \in \mathbb{R}^3$\\${\color{blue}\vf} \in \mathbb{R}^8$\\${\color{blue}\vs} \in \mathbb{R}^3$};

    \node[above=1pt of quantization,align=center,blue]{$\MLP_h\;,\;\mathrm{Hash}$};
    \node[above=2pt of render-0,align=center]{Rendered};
    \node[below=6pt of gt-0,align=center,xshift=4pt]{Ground truth};

    \draw[arrow,red,shorten >=0cm,shorten <=0.2cm,-] (render-0) -- node[left,red,yshift=-3pt]{} coordinate[midway] (redmid) (gt-0);
    \draw[arrow,red] ($(redmid)+(0pt,-3pt)$) -- ++(-1.0cm,0) node[left,red,xshift=5pt]{$(1-\lambda_\text{s})\mathcal{L}_1 + \lambda_\text{s}\mathcal{L}_\text{s}$};

  \end{tikzpicture}
  \end{minipage}}
  \caption{\textbf{Method overview of \ours:} We train (\textcolor{blue}{trainable parameters in blue}) neural splats with tiny MLP-decoders for view-dependent rendering (\cref{sec:neural_splats}). The coordinates are compressed with occupancy-octree coordinate coding (\cref{sec:coordinate-compression}). We also learn the quantization of splat features $\vf$ and scaling controllers $\vs$ with NLL rate terms (\cref{sec:feature-compression}), and employ adaptive density control with stage-wise training (\textcolor{red}{loss-terms in red}) to balance fidelity and size (\cref{sec:adc,sec:training_stages}).}
  \label{fig:main_architecture}
  \vspace*{-1.5em}
\end{figure*}

\ours consists of five parts that work together (see \cref{fig:main_architecture} for an overview):
{\em (i)}~\emph{Coordinate compression} (\cref{sec:coordinate-compression}): Splat coordinates are stored with an occupancy-octree and entropy coding, leveraging spatial sparsity while avoiding anchor-offset overhead. \sloppy 
{\em (ii)}~\emph{Positional encoding} (\cref{sec:pos_encoding}): Each splat inherits a unique identifier derived from the corresponding finest-level octree cell. This octree-derived positional encoding improves representation compactness and structural consistency.
{\em (iii)}~\emph{Tiny decoders} (\cref{sec:neural_splats}): Small MLPs reconstruct view-dependent color and transformation from abstract features, viewing direction, and positional embeddings, keeping rendering fast. 
{\em (iv)}~\emph{Feature compression} (\cref{sec:feature-compression}):
Per-splat features are quantized using learned step sizes and encoded arithmetically.
{\em (v)}~\emph{Adaptive density control and training} (\cref{sec:adc,sec:training_stages}): Gradient-based densification and pruning, together with stage-wise optimization, balance reconstruction fidelity and model size.\looseness-1

\subsection{Coordinate compression: Occupancy-octree encoding}
\label{sec:coordinate-compression}
Many 3DGS compression methods skip coordinate quantization due to the sensitivity of reconstruction to geometry. In \cref{fig:effect_of_recursion_quantitative}, we show that this precision requirement could be relaxed: 16-bit per-axis quantization yields an imperceptible quality drop. Therefore, we use an octree structure to capture the occupancy of splats in 3D space such that the tree is built recursively (see \cref{fig:occupancy_octree_encoding}), drawing intuition from \cite{schnabel2006octree,que2021voxelcontext,fu2022octattention,cui2023octformer}.

We begin from the axis-aligned bounding box covering all splats and, at recursion $r=1$, subdivide it into 8 equal sub-boxes. At each subsequent recursion, only \emph{non-empty} boxes (those containing at least one splat) are subdivided again into $8$ equal children. The boundary of the coarsest-level box is defined by the minimum and maximum of all splat coordinates: $\vx_{\min}$ and $\vx_{\max}$. A box at depth $r\in\mathbb{N}$ has size $b_r = (\vx_{\max}-\vx_{\min})/2^{r}$. Each subdivision emits an 8-bit \emph{occupancy byte} indicating which children are non-empty (1) or empty (0), with child order given by the Morton space-filling curve \cite{morton1966spacefilling}. After $R$ recursions, the resulting occupancy-octree explicitly represents the spatial layout of all splats such that the center of each finest-level box at depth $R$ provides the quantized coordinate, with at most one splat in each finest-level box. Excessive splats that fall into the same finest-level cell are pruned. This correspondence quantizes the coordinate of the splat $\vx$ to be 
\begin{equation}
  \label{eq:quantized_coor}
  \hat{\vx} = b_R \cdot \mathrm{floor}((\vx - \vx_{\min}) /b_R) + 0.5 b_R + \vx_{\min}.
\end{equation}
\textbf{Coordinate storage} then reduces to recording the sequence of occupancy bytes for all non-empty internal tree nodes, plus the boundaries of the coarsest bounding box  and recursion depth $R$. Because these bytes are highly sparse (most nodes have few occupied children, see \cref{app:octree_analysis}), we apply Huffman coding \cite{huffman2007method} for an entropy-based compression, yielding compact \emph{lossless-within-grid} storage of coordinates relative to the chosen quantization grid.

\subsection{Positional encoding: Index from space division}
\label{sec:pos_encoding}
During the spatial partitioning described in \cref{sec:coordinate-compression}, each splat is assigned to a unique finest-level bounding box. At each recursion, every box is subdivided into eight sub-boxes. Therefore, there are at most $8^{R}$ possible finest-level boxes available for assignment.
Each finest-level box can then be encoded by a $3R$-bit sequence, which serves as the \textbf{positional embedding} $\vf_{oct} \in \{0,1\}^{3R}$ of the splat contained within that box. Mathematically, $\vf_{oct}$ represents the binary data of the Morton code (Z-order index) \citep{morton1966spacefilling} of the associated finest-level box.

Suppose a splat is located in box $B$ at recursion depth $r-1$. Then, the $(3r-2)$\textsuperscript{th} through $3r$\textsuperscript{th} elements of $\vf_{oct}$ indicate which sub-box at recursion depth $r$, contained within $B$, includes the splat. For example, when there is only one recursion level, $\vf_{oct} = (1, 0, 0)$ indicates that the splat is located at the 4\textsuperscript{th} sub-box of the entire space, since `100' is the binary representation of `4'.

\subsection{Neural Gaussian splats}
\label{sec:neural_splats}
\ours models the 3D scene as a set of neural Gaussian splats with the support of a few MLPs ($\MLP_{o}$, $\MLP_{c}$, $\MLP_{s}$, $\MLP_{r}$, and $\MLP_{oct}$). Each splat has a coordinate $\vx\in\mathbb{R}^3$, a feature vector $\vf\in\mathbb{R}^{n_f}$ (we use $n_f = 8$ in practice, see \cref{sec:ablations}), and a scaling controller $\vs\in\mathbb{R}^3$. The feature $\vf$ encodes abstract, splat-wise cues including optical information, geometric transformation, and orientation-induced material properties.\looseness-1

\paragraph{Feature generation} Given the camera center $\vx_c$ during rendering, the augmented feature $\vf^*$ is constructed by concatenating the splat feature $\vf$ with viewing information and positional encoding,
\begin{equation}
  \label{eq:view_concat}
  \vf^* = \mathrm{concat}\left(\vf, \frac{\vx_c - \vx}{\|\vx_c - \vx\|}, \|\vx_c - \vx\|, \MLP_{oct}(\vf_{oct})\right),
\end{equation}
where $ \frac{\vx_c - \vx}{\|\vx_c - \vx\|}$ defines viewing direction,  $\|\vx_c - \vx\|$ is the camera-to-splat distance and $\vf_{oct}$ is the positional embedding defined in \cref{sec:pos_encoding}.
The per-splat optical attributes are then predicted as
\begin{equation}
  \begin{aligned}
  o &= |\MLP_{o}(\vf^*)|, &&\text{(opacity)} &
  \vc &= \MLP_{c}(\vf^*), &&\text{(color)} \\
  \vr &= \MLP_{r}(\vf^*), &&\text{(rotation)} &
  \vs^* &= \vs \odot \mathrm{sigmoid}(\MLP_{s}(\vf^*)), &&\text{(scaling)} \\
  \end{aligned} 
\end{equation}
where $o \in \mathbb{R}$, $\vc \in \mathbb{R}^3, \vr \in \mathbb{R}^4$, and $\vs^* \in \mathbb{R}^3$ denote the predicted opacity, color, rotation, and scaling, respectively, and $\odot$ denotes element-wise multiplication.  
Generating scaling parameters $\vs^*$ by relying solely on the MLP would cause instability and training collapses. To mitigate this issue, the parameter $\vs$ is introduced as a splat size bound, and the $\MLP_s$ predicts a modulation factor via a sigmoid activation. 
The scaling matrix $\mS$ and rotation matrix $\mR$ defined by $\vs^*$ and $\vr$ reconstruct the covariance $\bm{\Sigma}$ of the splat:
\begin{equation}
  \bm{\Sigma} = \mR \mS \mS^\top \mR^\top.
\end{equation}

\paragraph{Rasterization} The pixel-wise rasterization follows the $\alpha$-blending in the original 3DGS \citep{kerbl20233dgs}. For each pixel $\vx'\in \mathbb{R}^2$ in the 2D plane, the color $C(\vx')$ is obtained by projecting all neural splats onto the image plane and compositing them along the ray by blending:
\begin{equation}\textstyle
  C(\vx') = \sum_{i=1}^{N_\text{ray}} T_i o_i G_i(\vx') \vc_i, \quad T_i = \prod_{j=1}^{i-1} (1 - o_jG_j'(\vx')),
\end{equation}
where $N_\text{ray}$ is the number of splats along the ray where they are ordered by the distance to the camera center, $o_i$ and $\vc_i$ are the predicted splat opacity and color, respectively. $G(\vx')$ denotes the projected Gaussian weight:
\begin{equation}\textstyle
  G(\vx') = \exp\big(-\frac{1}{2}(\vx' - \bm{\mu}')^\top \bm{\Sigma}'^{-1} (\vx' - \bm{\mu}')\big).
\end{equation}
Here, $\bm{\mu}'$ and $\bm{\Sigma}' = \mJ \mW \bm{\Sigma}  \mW^\top\mJ^\top$ are the splat coordinate and covariance after projection onto the 2D plane \cite{kopanas2021point,yifan2019differentiable}, 
where $\mJ$ denotes the Jacobian of projective transformation and $\mW$ is the view matrix. 
Overall, the parametrization in \ours keeps the geometry explicit, view effects compact, and shares learned spatial information via the positional embeddings.

\begin{figure*}[t!]
  \centering\footnotesize
  \resizebox{\textwidth}{!}{
  \begin{minipage}{1.2\textwidth}
  \begin{tikzpicture}[scale=1]

  \definecolor{red}{RGB}{255,127,14}
  \definecolor{blue}{RGB}{31,119,180}
  
  \def\a{2}
  \pgfmathsetmacro{\h}{\a/2} %

    \tikzstyle{arrow} = [
      -{Stealth[length=3mm,width=2mm]},
      shorten >= 4pt,
      shorten <= -3pt,
      line width=0.9pt,
      draw=black!70,
      rounded corners=6pt
    ]

  \newcommand{\fillfaceXY}[4]{%
    \fill[#1] (0,0,\a) -- (\h,0,\a) -- (\h,\h,\a) -- (0,\h,\a) -- cycle;%
    \fill[#2] (\h,0,\a) -- (\a,0,\a) -- (\a,\h,\a) -- (\h,\h,\a) -- cycle;%
    \fill[#3] (0,\h,\a) -- (\h,\h,\a) -- (\h,\a,\a) -- (0,\a,\a) -- cycle;%
    \fill[#4] (\h,\h,\a) -- (\a,\h,\a) -- (\a,\a,\a) -- (\h,\a,\a) -- cycle;%
  }

  \newcommand{\fillfaceXZ}[4]{%
    \fill[#1] (0,\a,0) -- (\h,\a,0) -- (\h,\a,\h) -- (0,\a,\h) -- cycle;%
    \fill[#2] (\h,\a,0) -- (\a,\a,0) -- (\a,\a,\h) -- (\h,\a,\h) -- cycle;%
    \fill[#3] (0,\a,\h) -- (\h,\a,\h) -- (\h,\a,\a) -- (0,\a,\a) -- cycle;%
    \fill[#4] (\h,\a,\h) -- (\a,\a,\h) -- (\a,\a,\a) -- (\h,\a,\a) -- cycle;%
  }

  \newcommand{\fillfaceYZ}[4]{%
    \fill[#1] (\a,0,0) -- (\a,\h,0) -- (\a,\h,\h) -- (\a,0,\h) -- cycle;%
    \fill[#2] (\a,\h,0) -- (\a,\a,0) -- (\a,\a,\h) -- (\a,\h,\h) -- cycle;%
    \fill[#3] (\a,0,\h) -- (\a,\h,\h) -- (\a,\h,\a) -- (\a,0,\a) -- cycle;%
    \fill[#4] (\a,\h,\h) -- (\a,\a,\h) -- (\a,\a,\a) -- (\a,\h,\a) -- cycle;%
  }

  \tikzset{grid/.style={line width=0.3pt, draw=black!35}}
  \newcommand{\drawXYgrid}{%
    \draw[grid] (0,\h,\a)--(\a,\h,\a) ( \h,0,\a)--(\h,\a,\a);
  }
  \newcommand{\drawXZgrid}{%
    \draw[grid] (0,\a,\h)--(\a,\a,\h) ( \h,\a,0)--(\h,\a,\a);
  }
  \newcommand{\drawYZgrid}{%
    \draw[grid] (\a,0,\h)--(\a,\a,\h) (\a,\h,0)--(\a,\h,\a);
  }

  \begin{scope}

\begin{scope}[scale=1,transform shape]
\foreach \xyz in {(0.2,0.13,1.9),
(1.55, 0.25, 1.50),
(1.54, 0.53, 1.26),
(1.75, 0.17, 1.87),
(1.90, 0.01, 1.91),
(1.84, 0.12, 1.64),
(1.70, 0.56, 1.29),
(1.80, 0.32, 1.36),
(1.91, 0.55, 1.36),
(1.82, 0.19, 1.27),
(1.92, .43, 1.85),
(1.87, 0.10, 1.96),
(1.66, 0.50, 1.54),
(1.73, 0.54, 1.98),
(1.94, 0.46, 1.27),
(1.95, 0.13, 1.1),
(1.95, 0.07, 1.7),
(1.96, 0.85, 1.23),
(1.63, 0.36, 1.54),
(1.74, 0.39, 1.93),
(1.51, 0.20, 1.83),
(1.56, 0.29, 1.60),
(1.82, 0.58, 1.57),
(1.51, 0.33, 1.39),
(1.18, 0.18, 1.93),
(1.12, 0.37, 1.86),
(1.48, 0.17, 1.88),
(1.1, 0.85, 1.88),
(1.29, 0.46, 1.98),
(1.24, 0.67, 1.91),
(1.32, 0.43, 1.86),
(1.38, 0.79, 1.99),
(1.38, 0.74, 1.86),
(1.11, 0.39, 1.87),
(1.77, 1.99, 0.21),
(1.96, 1.85, 0.27),
(1.90, 1.90, 0.27),
(1.90, 1.97, 0.19),
(1.89, 1.98, 0.13)}
      \node[inner sep=1pt,circle,fill=black!80] at \xyz {};
\end{scope}
  
    \coordinate (O)   at (0,0,0);
    \coordinate (X)   at (\a,0,0);
    \coordinate (Y)   at (0,\a,0);
    \coordinate (Z)   at (0,0,\a);
    \coordinate (XY)  at (\a,\a,0);
    \coordinate (XZ)  at (\a,0,\a);
    \coordinate (YZ)  at (0,\a,\a);
    \coordinate (XYZ) at (\a,\a,\a);

    \draw[dashed,black!50] (O)--(X) (O)--(Y) (O)--(Z);

    \draw (XY)--(XYZ) (XZ)--(XYZ) (YZ)--(XYZ);
    \draw (X)--(XY) (X)--(XZ);
    \draw (Y)--(XY) (Y)--(YZ);
    \draw (Z)--(XZ) (Z)--(YZ); 

\node[align=center,font=\bf\strut] at ([yshift=1.25em]\a/2,\a,0) {Point cloud};
    
  \end{scope}

  \begin{scope}[xshift=.22\textwidth]

    \fillfaceXY{gray!50}{gray!50}{white}{white}
    \fillfaceXZ{white}{gray!50}{white}{white}
    \fillfaceYZ{white}{gray!50}{gray!50}{white}

    \tikzset{grid/.style={line width=0.5pt, draw=blue!80}}
    \drawXYgrid
    \drawXZgrid
    \drawYZgrid

    \coordinate (O)   at (0,0,0);
    \coordinate (X)   at (\a,0,0);
    \coordinate (Y)   at (0,\a,0);
    \coordinate (Z)   at (0,0,\a);
    \coordinate (XY)  at (\a,\a,0);
    \coordinate (XZ)  at (\a,0,\a);
    \coordinate (YZ)  at (0,\a,\a);
    \coordinate (XYZ) at (\a,\a,\a);

    \draw (XY)--(XYZ) (XZ)--(XYZ) (YZ)--(XYZ);
    \draw (X)--(XY) (X)--(XZ);
    \draw (Y)--(XY) (Y)--(YZ);
    \draw (Z)--(XZ) (Z)--(YZ);\

\begin{scope}[scale=1,transform shape]
\foreach \xyz in {(0.2,0.13,1.9),
(1.55, 0.25, 1.50),
(1.54, 0.53, 1.26),
(1.75, 0.17, 1.87),
(1.90, 0.01, 1.91),
(1.84, 0.12, 1.64),
(1.70, 0.56, 1.29),
(1.80, 0.32, 1.36),
(1.91, 0.55, 1.36),
(1.82, 0.19, 1.27),
(1.92, .43, 1.85),
(1.87, 0.10, 1.96),
(1.66, 0.50, 1.54),
(1.73, 0.54, 1.98),
(1.94, 0.46, 1.27),
(1.95, 0.13, 1.1),
(1.95, 0.07, 1.7),
(1.96, 0.85, 1.23),
(1.63, 0.36, 1.54),
(1.74, 0.39, 1.93),
(1.51, 0.20, 1.83),
(1.56, 0.29, 1.60),
(1.82, 0.58, 1.57),
(1.51, 0.33, 1.39),
(1.18, 0.18, 1.93),
(1.12, 0.37, 1.86),
(1.48, 0.17, 1.88),
(1.1, 0.85, 1.88),
(1.29, 0.46, 1.98),
(1.24, 0.67, 1.91),
(1.32, 0.43, 1.86),
(1.38, 0.79, 1.99),
(1.38, 0.74, 1.86),
(1.11, 0.39, 1.87),
(1.77, 1.99, 0.21),
(1.96, 1.85, 0.27),
(1.90, 1.90, 0.27),
(1.90, 1.97, 0.19),
(1.89, 1.98, 0.13)}
\node[inner sep=1pt,circle,fill=black!80] at \xyz {};
\end{scope}

\node (secondcube-left) at (0,\a/4,\a) {};
\node (secondcube-middle) at (1.5\a,0,\a) {};
\node (secondcube-right) at (\a,1.5\a,0) {};

\node[align=center,font=\bf\strut] at ([yshift=1.25em]\a/2,\a,0) {Occupancy division};
  \end{scope}

  \begin{scope}[xshift=.155\textwidth,yshift=-.14\textwidth,scale=.5]

    \fillfaceXY{gray!50}{white}{white}{white}
    \tikzset{grid/.style={line width=0.3pt, draw=red!80}}
    \drawXYgrid
    \drawXZgrid
    \drawYZgrid

    \coordinate (O)   at (0,0,0);
    \coordinate (X)   at (\a,0,0);
    \coordinate (Y)   at (0,\a,0);
    \coordinate (Z)   at (0,0,\a);
    \coordinate (XY)  at (\a,\a,0);
    \coordinate (XZ)  at (\a,0,\a);
    \coordinate (YZ)  at (0,\a,\a);
    \coordinate (XYZ) at (\a,\a,\a);

    \draw[blue!80] (XY)--(XYZ) (XZ)--(XYZ) (YZ)--(XYZ);
    \draw[blue!80] (X)--(XY) (X)--(XZ);
    \draw[blue!80] (Y)--(XY) (Y)--(YZ);
    \draw[blue!80] (Z)--(XZ) (Z)--(YZ);

    \node (leftsubcube) at (-.25\a,\a) {};

  \end{scope}

  \begin{scope}[xshift=.255\textwidth,yshift=-.14\textwidth,scale=.5]

    \fillfaceXY{gray!50}{gray!50}{gray!50}{gray!50}
    \fillfaceXZ{gray!50}{gray!50}{gray!50}{gray!50}
    \fillfaceYZ{gray!50}{gray!50}{gray!50}{gray!50}

    \tikzset{grid/.style={line width=0.3pt, draw=red!80}}
    \drawXYgrid
    \drawXZgrid
    \drawYZgrid

    \coordinate (O)   at (0,0,0);
    \coordinate (X)   at (\a,0,0);
    \coordinate (Y)   at (0,\a,0);
    \coordinate (Z)   at (0,0,\a);
    \coordinate (XY)  at (\a,\a,0);
    \coordinate (XZ)  at (\a,0,\a);
    \coordinate (YZ)  at (0,\a,\a);
    \coordinate (XYZ) at (\a,\a,\a);

    \draw[blue!80] (XY)--(XYZ) (XZ)--(XYZ) (YZ)--(XYZ);
    \draw[blue!80] (X)--(XY) (X)--(XZ);
    \draw[blue!80] (Y)--(XY) (Y)--(YZ);
    \draw[blue!80] (Z)--(XZ) (Z)--(YZ);  

\node (middlesubcube) at (\a/2,\a,0) {};

  \end{scope}

  \begin{scope}[xshift=.355\textwidth,yshift=-.14\textwidth,scale=.5]

    \fillfaceXZ{white}{gray!50}{white}{white}
    \fillfaceYZ{white}{gray!50}{white}{white}

    \tikzset{grid/.style={line width=0.3pt, draw=red!80}}
    \drawXYgrid
    \drawXZgrid
    \drawYZgrid

    \coordinate (O)   at (0,0,0);
    \coordinate (X)   at (\a,0,0);
    \coordinate (Y)   at (0,\a,0);
    \coordinate (Z)   at (0,0,\a);
    \coordinate (XY)  at (\a,\a,0);
    \coordinate (XZ)  at (\a,0,\a);
    \coordinate (YZ)  at (0,\a,\a);
    \coordinate (XYZ) at (\a,\a,\a);

    \draw[blue!80] (XY)--(XYZ) (XZ)--(XYZ) (YZ)--(XYZ);
    \draw[blue!80] (X)--(XY) (X)--(XZ);
    \draw[blue!80] (Y)--(XY) (Y)--(YZ);
    \draw[blue!80] (Z)--(XZ) (Z)--(YZ);  

    \node (rightsubcube) at (1.5\a,\a,0) {};

  \end{scope} 

\draw[arrow] ([xshift=2em,yshift=1em]secondcube-left) -| (leftsubcube);
\draw[arrow] ([xshift=0em,yshift=0em]secondcube-middle) -- ([xshift=0em,yshift=0em]middlesubcube);
\draw[arrow] ([xshift=0em,yshift=0em]secondcube-right) -| ([xshift=0em,yshift=0em]rightsubcube);

  \tikzstyle{leafA} = [
      draw=black!60,
      circle,
      minimum size=1.8em,
      thick
  ]
  \tikzstyle{leafB} = [
      draw=black!60,
      circle,
      minimum size=1.8em,
      thick
  ]
  \tikzstyle{leafC} = [
      draw=black!60,
      circle,
      minimum size=1em,
      thick
  ]

  \begin{scope}[xshift=.66\textwidth,yshift=0.1\textwidth,scale=1,transform shape]

    \node[leafA] (n0) at (2em,-0.65em) {};

    \foreach \bit [count=\i] in {1,0,1,0,0,0,0,1} {
      \ifnum\bit=1
        \node[leafB,blue!80] (n0\i) at ($(n0)+(-11em,-4.7em)+(2.5em*\i,0)$) {};
        \draw[black!60] (n0) -- (n0\i);
      \else
        \node[leafB,blue!80,densely dotted] (n0\i) at ($(n0)+(-11em,-4.7em)+(2.5em*\i,0)$) {};
        \draw[black!60,densely dotted] (n0) -- (n0\i);
      \fi
      \node[anchor=north,inner sep=0,font=\strut\footnotesize] at (n0\i.south) {\bit};
    }

    \foreach \bit [count=\i] in {1,0,0,0,0,0,0,0} {
     \ifnum\bit=1
        \node[leafC,red!80] (n01\i) at ($(n01)+(-7em,-5em)+(1.2em*\i,0)$) {};
        \draw[black!60] (n01) -- (n01\i);
      \else
        \node[leafC,red!80,densely dotted] (n01\i) at ($(n01)+(-7em,-5em)+(1.2em*\i,0)$) {};
        \draw[black!60,densely dotted] (n01) -- (n01\i);
      \fi
      \node[anchor=north,inner sep=0,font=\strut\tiny] at (n01\i.south) {\bit};
    }

    \foreach \bit [count=\i] in {1,1,1,1,1,1,1,1} {
     \ifnum\bit=1
        \node[leafC,red!80] (n03\i) at ($(n03)+(-1.3em,-5em)+(1.2em*\i,0)$) {};
        \draw[black!60] (n03) -- (n03\i);
      \else
        \node[leafC,red!80,densely dotted] (n03\i) at ($(n03)+(-1.3em,-5em)+(1.2em*\i,0)$) {};
        \draw[black!60,densely dotted] (n03) -- (n03\i);
      \fi
      \node[anchor=north,inner sep=0,font=\strut\tiny] at (n03\i.south) {\bit};
    }

    \foreach \bit [count=\i] in {0,0,0,0,0,0,0,1} {
     \ifnum\bit=1
        \node[leafC,red!80] (n08\i) at ($(n08)+(-3em,-5em)+(1.2em*\i,0)$) {};
        \draw[black!60] (n08) -- (n08\i);
      \else
        \node[leafC,red!80,densely dotted] (n08\i) at ($(n08)+(-3em,-5em)+(1.2em*\i,0)$) {};
        \draw[black!60,densely dotted] (n08) -- (n08\i);
      \fi
      \node[anchor=north,inner sep=0,font=\strut\tiny] at (n08\i.south) {\bit};
    }

\node[align=center,font=\bf\strut] at ([yshift=1.25em]n0.north) {Occupancy-octree};
\node[align=center, font=\small,color=blue!80] at ([xshift=2.5em]n08.east) {$C_{11}$};
\node[align=center, font=\small,color=red!80] at ([xshift=0.7em,yshift=-1.7em]n014.south) {$C_{21}$};
\node[align=center, font=\small,color=red!80] at ([xshift=0.7em,yshift=-1.7em]n034.south) {$C_{23}$};
\node[align=center, font=\small,color=red!80] at ([xshift=0.7em,yshift=-1.7em]n084.south) {$C_{28}$};

\node (tree-base) [align=left, font=\small] at ([xshift=-7em,yshift=-13.5em]n0.south) {$\left[C_{11}, C_{21}, C_{23}, C_{28},\cdots \right]$};

\draw[->] ($(tree-base)+(51pt,0pt)$) -- ++(2.2cm,0cm) node[midway, above,xshift=-2pt,yshift=0pt,font=\scriptsize]{Huffman coding};
\node[align=left] at ([xshift=5.3cm]tree-base) {\texttt{1100101001...}};

  \end{scope}

  \end{tikzpicture}
  \end{minipage}}
  \vspace{.5em}
  \caption{\textbf{Occupancy-octree coordinate coding:} Given a point cloud (left), we recursively divide the bounding box into eight sub-boxes. Only the non-empty sub-boxes (gray) are further divided (middle). Each division is represented by an 8-bit binary code (1=non-empty, 0=empty). $C_{ri}$ denotes the code of the $i$\textsuperscript{th} division at $r$\textsuperscript{th} recursion. The occupancy octree (right) is encoded by arranging all bits in a breadth-first manner. Finally, we apply Huffman coding to further compress the octree bits.}
  \label{fig:occupancy_octree_encoding}
  \vspace{-1em}
\end{figure*}

\subsection{Splat-wise feature compression: Arithmetic coding}
\label{sec:feature-compression}

Each splat carries the scaling controller $\vs\in\mathbb{R}^3$ and an abstract feature $\vf\in\mathbb{R}^{n_f}$ encoding color, opacity, transformation, and material cues (see \cref{sec:neural_splats}). We compress them by hash-grid assisted arithmetic coding and a Gaussian distribution prior~\cite{chen2024hac}.
\ours uses low-dimensional features $n_f = 8$ (more details in \cref{app:ablation_study}), much smaller than the 50 dimensions in most anchor-based methods~\cite{chen2025hac++, lu2024scaffoldgs}.

First, we condition feature coding on spatial hash descriptors by predicting a distribution and a per-dimension step size with a tiny network $\MLP_h$:
  $\bm{\mu}_\vf, \bm{\Sigma}_\vf, \Delta_f = \MLP_{h}(\mathrm{Hash}(\vx))$,
where $\mathrm{Hash}(\cdot)$ denotes the hash encoding function of InstantNGP~\cite{muller2022instantngp}. For better compression, the parameters of $\mathrm{Hash}(\cdot)$ are quantized as binary. We then quantize the feature $\vf$ with the learned steps,
  $\hat{\vf} = \Delta_f \odot \mathrm{round}(\vf / \Delta_f)$,
and assume the quantized features follow a Gaussian distribution of $\mathcal{N}(\vf; \bm{\mu}_\vf, \bm{\Sigma}_\vf)$ (with diagonal $\bm{\Sigma}_\vf$). The negative log-likelihood (NLL) of the quantization bin provides a rate proxy:
\begin{equation}\textstyle
  \mathrm{NLL}(\vf) = -\log \int_{\hat{\vf}-\Delta_f}^{\hat{\vf}+\Delta_f} \mathcal{N}(\vf'; \bm{\mu}_\vf, \bm{\Sigma}_\vf) \,\mathrm{d}\vf'.
\end{equation}
The NLL is the theoretical lower bound for the number of bits needed to encode the features. The average NLL over all $N$ splats serves as the training loss to encourage reducing the storage for encoding the features:
  $\mathcal{L}_\text{NLL}(\vf) = \frac{1}{N} \sum_{i=1}^{N} \mathrm{NLL}(\vf_i)$.
We use arithmetic coding \cite{witten1987arithmetic} to compress both features $\vf$ and the scaling controller $\vs$ of the splats. The loss function $\mathcal{L}_\text{NLL}(\vs)$ and quantization $\hat{\vs}$ of the scaling controller $\vs$ are defined similarly to the above.

\subsection{Adaptive density control}
\label{sec:adc}
\ours employs an adaptive density control (ADC) strategy to maintain high-fidelity while keeping a sparse distribution of splats.
\textbf{Initialization:}
Following 3DGS~\cite{kerbl20233dgs}, we initialize coordinates $\vx$ and scaling controllers $\vs$ from SfM point clouds \cite{schonberger2016sfm}. Because SfM points are sparse and non-uniformly scattered, some regions require densification. The features $\vf$ are initialized randomly.
\textbf{Densification:} 
Splats with a high magnitude of the coordinate gradient are densified during training~\cite{lu2024scaffoldgs}. We choose threshold $\tau_g$ such that the splat whose coordinate's gradient magnitude is higher than $\tau_{g}$ will be cloned as a new splat. The newly introduced splat inherits the feature $\vf$ from the parent splat, while its coordinate is randomly sampled from a Gaussian $\mathcal{N}(\vx, \bm{\Sigma})$, where $\vx$ and $\bm{\Sigma}$ are the coordinate and covariance of the parent splat, respectively. This process defines new splats and adds detail to under-represented regions of the scene, improving reconstruction quality while increasing overall density.
\textbf{Pruning:} 
During training, an individual splat is pruned when either its average opacity is less than the threshold $\tau_{o}$, or if it remains consistently invisible. %
\textbf{Opacity regularization:} 
To encourage sparsity in the splat distribution, we apply an $L_1$ regularization loss on the opacity of all $N$ splats during specific stages of training (see \cref{sec:training_stages}): 
  $\mathcal{L}_{o} = \sum_{i=1}^{N} |\MLP_o(\vf_i^*)|.$

\subsection{Training pipeline for \ours}
\label{sec:training_stages}
We provide a consistent training pipeline, where \ours is trained with the following five stages: 
{\em (1)}~{\em Warm-up stage} (0--0.5k): The splat coordinates $\vx$ and scaling controllers $\vs$ are initialized from SfM, and abstract features $\vf$ are randomly initialized. 
{\em (2)}~{\em Densification stage} (0.5k--15k): The number of splats is controlled mutually by densification and pruning as described in \cref{sec:adc}. 
{\em (3)}~{\em Compaction stage} (15k--20k): Both opacity regularization and pruning modules further reduce the number of splats. 
{\em (4)}~{\em Feature compression stage} (20k--30k): Both features $\vf$ and scaling controllers $\vs$ of the splats are quantized and compressed via arithmetic coding. 
{\em (5)}~{\em Coordinate compression stage} (30k--35k): All splat coordinates are quantized and stored by the occupancy-octree technique. 
In \cref{fig:training_curves}, we provide training curves for an example scene ({\it Garden}), which gives an idea of the influence of the different stages. Note that the proposed training schedule is not tuned on a per-scene or per-dataset basis.

\begin{figure}[t!]
  \centering\tiny
  \setlength{\figurewidth}{.75\textwidth}
  \setlength{\figureheight}{.35\textwidth}
  \pgfplotsset{
    xtick={0},
    xticklabels={0},
    scaled ticks=false,
    ylabel near ticks
  }
  \input{\figpath/training_curve_x3.tex}\\
  \vspace*{-.1in}
  \caption{\textbf{Training curves over different stages:} We log the number of splats (left axis), PSNR (middle), and compressed model size (right) over different stages of training, showing how the compaction and compression do not degrade the overall quality. The number of splats increases steadily during densification (0.5k--15k), then decreases rapidly during compaction (15k--20k).}
  \label{fig:training_curves}
  \vspace*{-1.5em}
\end{figure}

\paragraph{Optimization objective} 
We optimize a photometric objective combining the $L_1$ loss $\mathcal{L}_1$ and SSIM loss $\mathcal{L}_\text{s}$ between the rendered image and the ground truth, the opacity regularization loss $\mathcal{L}_{o}$, and the quantization regularization losses of $\mathcal{L}_\text{NLL}(\vf)+ \mathcal{L}_\text{NLL}(\vs)$ such that
\begin{equation}
  \label{eq:training_loss}
\mathcal{L} = (1-\lambda_\text{s}) \mathcal{L}_1 + \lambda_\text{s} \mathcal{L}_\text{s} + \lambda_{o} \mathcal{L}_{o} + \lambda_{q}\left[\mathcal{L}_\text{NLL}(\vf)+ \mathcal{L}_\text{NLL}(\vs)\right].
\end{equation}
Here $\lambda_\text{s}, \lambda_{o}$, and $\lambda_{q}$ are hyperparameters to control the optimization priorities of the SSIM loss, opacity regularization, and quantization regularization. The opacity regularization is only effective (\ie, $\lambda_{o}>0$) in the compaction stage {\em (3)}, and the quantization regularization is only active (\ie, $\lambda_{q}>0$) during the compression stages of {\em (4)} and {\em (5)}.

\section{Experiments}
\label{sec:experiments}
We evaluate \ours as a full 3DGS compression pipeline. Concretely, we demonstrate our rendering quality vs.\ model size both quantitatively and qualitatively under the well-known 3DGS.zip survey benchmark setup \cite{bagdasarian20253dgszip}. Additionally, we demonstrate storage size per module, analyze the inference speed, and ablate central parts of our model.

\paragraph{Data sets} We benchmark our \ours on standard permissively-licensed real-world 3D reconstruction data sets: nine scenes from \mipnerf \cite{ds_MipNeRF360}, two scenes from \tandt \cite{ds_tanks_and_temples}, and two scenes from \deepblending \cite{ds_deep_blending}. 
The training--testing split follows the original settings of 3DGS \cite{kerbl20233dgs} and Mip-NeRF~360 \citep{ds_MipNeRF360}: Every 8\textsuperscript{th} image is used for testing, and the rest are for training.

\paragraph{Evaluation metrics}
The reconstruction quality is evaluated by three widely used metrics: the Peak Signal-to-Noise Ratio (PSNR), the Structural Similarity Index Measure (SSIM), and the Learned Perceptual Image Patch Similarity (LPIPS) \cite{Zhang_2018_CVPR_lpips}. The compression is measured by the final model size in megabytes (MB). The reported reconstruction quality metrics represent the average value computed between the ground-truth images (from test sets) and the rendered images from the model.

\begin{wraptable}{r}{0.55\textwidth}
  \vspace*{-1em}
  \centering\scriptsize
  \caption{The size of components of \ours-base (in MB), averaged over all scenes.
  }
  \label{table:size_of_components}
  \setlength{\tabcolsep}{3.3pt}
\begin{tabular}{lc|ccccc}
\toprule
 & Total & ${\vx}$ & ${\vf}$ & ${\vs}$ & MLPs & Others \\
\midrule
\mipnerf & 5.860 & 1.186 & 2.753 & 1.751 & 0.157 & 0.013 \\
\deepblending & 2.860 & 0.739 & 1.097 & 0.846 & 0.169 & 0.009 \\
\tandt & 4.784 & 0.843 & 2.260 & 1.526 & 0.143 & 0.012 \\
\bottomrule
\end{tabular}
   \vspace*{-1em}
\end{wraptable}

\paragraph{Storage}
\ours is stored as multiple components: {\em (i)}~occupancy-octree data as bit strings to represent quantized coordinates $\hat{\vx}$, {\em (ii)}~binary arithmetic codes representing compressed features $\hat{\vf}$ and scaling controllers $\hat{\vs}$, {\em (iii)}~MLP architecture/weights ($\MLP_{o}$, $\MLP_{c}$, $\MLP_{s}$, $\MLP_{r}$, $\MLP_{oct}$, $\MLP_{h}$), and {\em (iv)}~additional metadata (\ie, the lower and upper bounds of the scene, the recursion depth $R$ and $Hash(\cdot)$). The total \ours size reported in this paper is the size of a single zip file containing all these four components (following the 3DGS.zip survey \cite{bagdasarian20253dgszip}). \cref{table:size_of_components} showns the size of each component in \ours averaged over scenes.

\paragraph{Implementation details} 
\ours is implemented in PyTorch \cite{paszke2019pytorch} using the Inria 3DGS repository \cite{kerbl20233dgs} as a starting point. Parameters of the model are trained using the Adam optimizer \cite{kingma2014adam}. The loss coefficients in \cref{eq:training_loss} are set to $\lambda_{s}=0.2$, and $\lambda_{q}=5\times 10^{-4}$.  The thresholds for densification and pruning are set to $\tau_{g} = 2 \times 10^{-4}$ and $\tau_{o} = 0.005$. We fix the feature $\vf$ dimensionality to $n_f=8$, and all scenes are discretized with $R=16$ recursion depth for coordinate compression and positional encoding. Additional implementation details, including learning rate (lr) values, lr schedulers, and MLP architectures, are provided in \cref{app:implementation_details}.
\textbf{\ours variants:} We include three variants of \ours (small, base, large) in the results with different compaction hyperparameters $\lambda_o$ (small: $\lambda_{o}=4{\times}10^{-7}$, base: $\lambda_{o}=3{\times}10^{-7}$, large: $\lambda_{o}=1{\times}10^{-7}$).
\textbf{Baselines:} We compare \ours with the most efficient and recent 3DGS compression methods of Compact3DGS~\cite{lee2024compact3d}, SOG~\cite{morgenstern2024sog},  Reducing3DGS~\cite{papantonakis2024reducing3d}, Compressed3D~\cite{niedermayr2024compressed3d}, CompGS~\cite{navaneet2024compgs}, HAC~\cite{chen2024hac}, ContextGS~\cite{wang2024contextgs}, HEMGS~\cite{liu2024hemgs}, and HAC++~\cite{chen2025hac++}.  We also provide the original vanilla 3DGS results as the prototype baseline.

\begin{figure*}[t!]
  \footnotesize\centering
  \captionof{table}{\textbf{\ours provides state-of-the-art quality--size trade-off:} We report PSNR/SSIM/LPIPS and the \emph{total model size in MBs} following the 3DGS.zip survey benchmark \cite{bagdasarian20253dgszip}. Across \mipnerf, \tandt, and \deepblending, \ours has the smallest sizes with better or similar quality to the strongest baselines, indicating a better rate--distortion trade-off.} %
  \label{table:quantitative_benchmark}
\newcommand{\MBMin}{2.5}   %
\newcommand{\MBMax}{50}  %
\newcommand{\mb}[2]{%
  \begingroup
  \pgfmathsetmacro{\v}{#2}%
  \pgfmathsetmacro{\den}{ln(\MBMax)/ln(2)-ln(\MBMin)/ln(2)}%
  \pgfmathsetmacro{\traw}{\den==0 ? 0 : (ln(\v)/ln(2)-ln(\MBMin)/ln(2))/\den}%
  \pgfmathsetmacro{\tclamp}{min(1,max(0,\traw))}%
  \pgfmathparse{round(100*\tclamp)}%
  \edef\X{\pgfmathresult}%
  \edef\tempcolor{white!\X!green}%
  \expandafter\cellcolor\expandafter{\tempcolor}{#1#2}%
  \endgroup
}  
  \setlength{\tabcolsep}{5pt}
  \newcommand{\nan}{---}
  \newcommand{\boldvalue}[1]{\bf#1}
  \newcommand{\rotate}[2]{\parbox[c]{2mm}{\multirow{#1}{*}{\tikz[baseline]\node[anchor=base,rotate=90,font=\tiny,align=center]{#2};}}}
  \resizebox{\textwidth}{!}{%
\begin{tabular}{ll|cccc|cccc|cccc}
    \toprule
    && \multicolumn{4}{c}{\mipnerf} & \multicolumn{4}{c}{\tandt} & \multicolumn{4}{c}{\deepblending}
    \\ 
    &Methods & PSNR$\uparrow$ & SSIM$\uparrow$ & LPIPS$\downarrow$ & Size$\downarrow$ &PSNR$\uparrow$ & SSIM$\uparrow$ & LPIPS$\downarrow$ & Size$\downarrow$ & PSNR$\uparrow$ & SSIM$\uparrow$ & LPIPS$\downarrow$ & Size$\downarrow$
    \\\midrule
&3DGS-30K \citep{kerbl20233dgs} & 27.21 & 0.815 & 0.214 & 734.000 & 23.14 & 0.841 & 0.183 & 411.000 & 29.41 & 0.903 & 0.243 & 676.000 \\
\midrule
&Compact3DGS \citep{lee2024compact3d}& 27.08 & 0.798 & 0.247 & \mb{}{48.800} & 23.32 & 0.831 & 0.201 & \mb{}{39.400} & 29.79 & 0.901 & 0.258 & \mb{}{43.200} \\
&SOG \citep{morgenstern2024sog}& 27.08 & 0.799 & 0.230 & \mb{}{40.285} & 23.56 & 0.837 & \boldvalue{0.186} & \mb{}{22.779} & 29.26 & 0.894 & 0.268 & \mb{}{17.737} \\
&Reducing3DGS \citep{papantonakis2024reducing3d}& 27.10 & \boldvalue{0.809} & \boldvalue{0.226} & \mb{}{29.000} & 23.57 & 0.840 & 0.188 & \mb{}{14.000} & 29.63 & 0.902 & \boldvalue{0.249} & \mb{}{18.000} \\
&Compressed3D \citep{niedermayr2024compressed3d}& 26.98 & 0.801 & 0.238 & \mb{}{28.803} & 23.32 & 0.832 & 0.194 & \mb{}{17.282} & 29.38 & 0.898 & 0.253 & \mb{}{25.299} \\
&CompGS \citep{navaneet2024compgs}& 27.03 & 0.804 & 0.243 & \mb{}{18.000} & 23.39 & 0.836 & 0.200 & \mb{}{12.000} & 29.90 & 0.906 & 0.252 & \mb{}{12.000} \\
&HAC \citep{chen2024hac}& 27.53 & 0.807 & 0.238 & \mb{}{16.005} & 24.04 & 0.846 & 0.187 & \mb{}{8.494} & 29.98 & 0.902 & 0.269 & \mb{}{4.561} \\
&ContextGS \citep{wang2024contextgs}& 27.62 & 0.808 & 0.237 & \mb{}{13.297} & 24.12 & \boldvalue{0.849} & \boldvalue{0.186} & \mb{}{9.902} & 30.09 & 0.907 & 0.265 & \mb{}{3.655} \\
&HEMGS \citep{liu2024hemgs}& 27.75 & 0.806 & 0.248 & \mb{}{12.539} & \boldvalue{24.42} & 0.848 & 0.192 & \mb{}{6.034} & \boldvalue{30.24} & \boldvalue{0.908} & 0.266 & \mb{}{2.993} \\
&HAC++ \citep{chen2025hac++}& 27.60 & 0.803 & 0.253 & \mb{}{8.742} & 24.22 & \boldvalue{0.849} & 0.190 & \mb{}{5.427} & 30.16 & 0.907 & 0.266 & \mb{}{3.051} \\
&Smol-GS-base & \boldvalue{27.76} & 0.807 & 0.248 & \mb{}{5.860} & 24.25 & 0.843 & 0.199 & \mb{}{4.784} & 30.12 & 0.905 & 0.262 & \mb{}{2.860} \\
&Smol-GS-small & 27.61 & 0.801 & 0.258 & \mb{\boldvalue}{4.867} & 24.21 & 0.841 & 0.204 & \mb{\boldvalue}{4.209} & 29.97 & 0.903 & 0.266 & \mb{\boldvalue}{2.422} \\
\bottomrule
    \end{tabular} %
}
  \vspace*{8pt}
\centering\small
  \setlength{\figurewidth}{.7\textwidth}
  \setlength{\figureheight}{.7\textwidth}
  \pgfplotsset{
    y tick label style={rotate=90},
    semithick/.append style={line width=3pt,opacity=.8},
    legend style={font=\footnotesize},
    ylabel style={font=\large},
    xlabel style={font=\large},
    grid style={densely dotted,lightgray!50},
    legend style={draw=none,inner xsep=2pt, inner ysep=0.5pt, nodes={inner sep=1.5pt, text depth=0.1em,scale=1,transform shape},fill=white,fill opacity=0.8}
    }
    \foreach \metric [count=\i] in {PSNR, SSIM, LPIPS} {
      \begin{subfigure}[b]{0.31\textwidth}
        \raggedleft 
        \scalebox{.45}{\input{\figpath/quantitative_scatter_plot/MipNeRF360_\metric.tex}}
      \end{subfigure}
    }
  \vspace*{0pt}
  \definecolor{steelblue31119180}{RGB}{31,119,180}
  \definecolor{lightgrey}{RGB}{211,211,211}
  \captionof{figure}{\textbf{Comparison of top methods on {\bfseries\scshape \mipnerf}.} Each curve refers to one method of different variants; the $x$-axis is the model size in MBs (smaller is better) and the $y$-axis reports PSNR/SSIM (higher is better) or LPIPS (lower is better). \ours (small--base--large) consistently lies beyond the Pareto frontier, achieving markedly smaller sizes at better quality to strong baselines collected via the 3DGS.zip benchmark. See \cref{fig:quantitative_benchmark_scatter} in the Appendix for results on \tandt and \deepblending.}
  \label{fig:quantitative}
  \vspace*{-1em}
\end{figure*}

\subsection{Results on 3DGS compression benchmarks}
\textbf{Quantitative results} We compare \ours with the baselines on three benchmark data sets using the four evaluation metrics mentioned earlier. The quantitative results are summarized in \cref{table:quantitative_benchmark} and \cref{fig:quantitative}.  The baseline results are collected from the \emph{3DGS.zip} compression survey \citep{bagdasarian20253dgszip}. In \cref{table:quantitative_benchmark}, \ours achieves the highest compression ratio of all baselines with a considerable margin, reducing the model size by about 150$\times$ compared to vanilla 3DGS while maintaining high reconstruction fidelity. In the scatter plots in \cref{fig:quantitative}, \ours outperforms all current state-of-the-art baselines in terms of the trade-off between compression ratio and reconstruction quality. 
We include further scatter plots for \tandt and \deepblending data sets in \cref{fig:quantitative_benchmark_scatter} in the Appendix.

\paragraph{Qualitative results} \cref{fig:qualitative} (\mipnerf) and \cref{fig:qualitative-tandt-db} (\tandt and \deepblending) show qualitative comparison of \ours-base, vanilla 3DGS, and the ground-truth (GT) test views for different scenes. We highlight specific regions by providing zoomed-in views for better subjective comparison, especially for challenging objects such as the white wall in the {\it Bonsai} and {\it Playroom} scenes, the sky in the {\it Train} scene, and the rocket sticker in the {\it Playroom} scene.

\begin{wraptable}{r}{0.45\textwidth}
  \centering\scriptsize
  \captionof{table}{The comparison of frames per second (FPS) for rendering among \ours-base, vanilla 3DGS, and HAC++. The values are averaged over all scenes.}
  \label{table:fps_comparison}
  \vspace*{-4pt}
  \setlength{\tabcolsep}{5pt}
\begin{tabular}{l|ccc}
\toprule
 & \ours & 3DGS & HAC++ \\
\midrule
\mipnerf & 210.0 & 139.3 & 128.0 \\
\deepblending & 399.9 & 178.8 & 149.8 \\
\tandt & 223.3 & 218.9 & 102.7 \\
\bottomrule
\end{tabular}
 
  \vspace*{1em}

  \centering\scriptsize
   \vspace*{0em}
  \captionof{table}{Ablation on different modules for compression. We report values on \mipnerf for \ours-base and its ablated variants: without Huffman coding, coordinate and feature compressions respectively. The rendered test views show that \ours is capable of reconstructing scene details with high fidelity and preserving sharp edges.
  }
  \label{table:ablation_compression_modules}
  \setlength{\tabcolsep}{3pt}
\begin{tabular}{l|cccc}
\toprule
Ablation & Size & PSNR $\uparrow$ & SSIM $\uparrow$ & LPIPS $\downarrow$ \\
\midrule
\ours-base & 5.86 & 27.76 & 0.807 & 0.248 \\
w/o Huffman & 7.038 & 27.76 & 0.807 & 0.248 \\
w/o Coordinate Quant. & 10.79 & 27.74 & 0.809 & 0.246 \\
w/o Feature Quant. & 23.71 & 27.72 & 0.805 & 0.250 \\
\bottomrule
\end{tabular} 
  \vspace*{-2em}

  \end{wraptable}

\ours clearly generates fewer artefacts than vanilla 3DGS. We link these visual improvements to the neural representation of abstract splats, which is capable of learning the optical properties and material cues of different surfaces adaptively. This representation is substantially more efficient in storing view-dependent cues compared to the expensive-to-store 3\textsuperscript{rd}-degree spherical harmonic coefficients in 3DGS.\looseness-1

\paragraph{Storage size}
The required disk space for storing each model component in \ours-base is summarized in \cref{table:size_of_components}. The splat-wise features $\vf$ take half of the space, while storing the coordinates $\vx$ is very efficient thanks to the occupancy-octree encoding. The MLP parameters solely take a negligible portion of the total size due to their tiny architectures. Compared to other methods like HAC++, we save more storage by eliminating the anchor-offset overhead and using low-dimensional features ($n_f=8$).

\paragraph{Inference speed}
The rendering speed of \ours-base surpasses both vanilla 3DGS and the anchor-based HAC++, as shown in \cref{table:fps_comparison}. Compared with 3DGS, the improvement stems from the view-dependent neural representation of splats, which is more storage-efficient than the costly 3\textsuperscript{rd}-degree spherical harmonic coefficients. Compared with HAC++, the gain arises from removing the anchor-offset representation, avoiding the additional complications associated with offset localization and transparent offsets.
The high rendering speed further demonstrates the practicality of \ours for real-time applications.
\begin{figure*}[t!]
    \centering\tiny
    \setlength{\figureheight}{.111\textwidth}        
    \setlength{\figurewidth}{1.5\figureheight}
    \setlength{\spysize}{0.48\figureheight}
    \foreach \method/\methodname [count=\i from 0] in {gt/{\strut GT},3dgs/{\strut 3DGS-30K},smolgs-base/{\strut \ours}} {        

    \begin{tikzpicture}[
        inner sep=0,
        image/.style = {inner sep=0pt, outer sep=1pt, , anchor=north west}, %
        node distance = 1pt and 1pt, every node/.style={font= {\tiny}}, 
        label/.style = {font={\footnotesize\bf\strut},anchor=south,inner sep=0pt}, 
        spy using outlines={rectangle, magnification=2, size=\spysize}
        ] 

    \foreach \scene/\scenename [count=\j from 0] in {train/Train,truck/Truck,drjohnson/Dr~Johnson,playroom/Playroom} {
      \ifnum\j=0
        \node [image] (img-\i\j) at (0,-\i*\figureheight) 
          {\includegraphics[height=\figureheight]{\figpath/qualitative_comparison/\method_\scene}};
      \else
        \pgfmathtruncatemacro{\prev}{\j-1}
        \node [image] (img-\i\j) at ($(img-\i\prev.north east)+(\spysize,0)$)
          {\includegraphics[height=\figureheight]{\figpath/qualitative_comparison/\method_\scene}};
      \fi

      \ifnum\i=0
        \node[label,anchor=south,font=\strut\bf] at (img-0\j.north) {\itshape \scenename};
      \fi

      \ifnum\j=0
        \node[label,rotate=90,anchor=south,font=\strut] at (img-\i0.west) {\bf \methodname};
      \fi

      \coordinate (spypoint-\j-a) at ($(img-\i\j.north east) + (0,-0.02\figurewidth)$);
      \coordinate (spypoint-\j-b) at ($(img-\i\j.south east) + (0,+0.02\figurewidth)$);       
      
    }

    \coordinate (spot-1-a) 
      at ($(img-\i0.south west) + (1.12\figurewidth,.5\figurewidth)$); %
    \spy[blue,magnification=2] on 
      (spot-1-a) in node[anchor=north west] at (spypoint-0-a);
    \coordinate (spot-1-b) 
      at ($(img-\i0.south west) + (.37\figurewidth,.57\figurewidth)$); %
    \spy[red,magnification=4] on 
      (spot-1-b) in node[anchor=south west] at (spypoint-0-b);

    \coordinate (spot-2-a) 
      at ($(img-\i1.south west) + (0.09\figurewidth,.5\figurewidth)$); %
    \spy[blue,magnification=4] on 
      (spot-2-a) in node[anchor=north west] at (spypoint-1-a);
    \coordinate (spot-2-b) 
      at ($(img-\i1.south west) + (0.27\figurewidth,.15\figurewidth)$); %
    \spy[red,magnification=1.5] on 
      (spot-2-b) in node[anchor=south west] at (spypoint-1-b);  

    \coordinate (spot-3-a) 
      at ($(img-\i2.south west) + (0.05\figurewidth,.4\figurewidth)$); %
    \spy[blue,magnification=4] on 
      (spot-3-a) in node[anchor=north west] at (spypoint-2-a);
    \coordinate (spot-3-b) 
      at ($(img-\i2.south west) + (0.92\figurewidth,.58\figurewidth)$); %
    \spy[red,magnification=1.5] on 
      (spot-3-b) in node[anchor=south west] at (spypoint-2-b);   

    \coordinate (spot-4-a) 
      at ($(img-\i3.south west) + (0.95\figurewidth,.6\figurewidth)$); %
    \spy[blue,magnification=4] on 
      (spot-4-a) in node[anchor=north west] at (spypoint-3-a);
    \coordinate (spot-4-b) 
      at ($(img-\i3.south west) + (0.8\figurewidth,.6\figurewidth)$); %
    \spy[red,magnification=2] on 
      (spot-4-b) in node[anchor=south west] at (spypoint-3-b);  

    \end{tikzpicture}\vspace{0.1em}
    }
    \caption{\textbf{Qualitative results on the {\bfseries\scshape Tanks and Temples} and {\bfseries\scshape Deep Blending} data sets.} \ours has fewer artefacts and better preservation of scene details under challenging lighting and materials. The learned splat features in \ours capture view-dependent and material cues more compactly than SH, aligning with its smaller size in MBs and competitive metrics in \cref{table:quantitative_benchmark}.}
    \label{fig:qualitative-tandt-db}
    \vspace*{-1em}
\end{figure*}

\begin{wrapfigure}{r}{0.45\textwidth}
  \raggedleft
  \vspace*{-1em}
  \setlength{\figurewidth}{.65\textwidth}
  \setlength{\figureheight}{.55\textwidth}
  \pgfplotsset{
    y tick label style={rotate=90},
    semithick/.append style={line width=3pt,opacity=.8},
    legend style={font=\large},
    ylabel style={font=\large},
    xlabel style={font=\large},
    grid style={densely dotted,lightgray!50},
    legend style={draw=none,inner xsep=2pt, inner ysep=0.5pt, nodes={inner sep=1.5pt, text depth=0.1em,scale=1,transform shape},fill=white,fill opacity=0.8}
    }
  \resizebox{.43\textwidth}{!}{
\begin{tikzpicture}

\definecolor{darkgray176}{RGB}{176,176,176}
\definecolor{darkorange25512714}{RGB}{255,127,14}
\definecolor{lightgray204}{RGB}{204,204,204}
\definecolor{steelblue31119180}{RGB}{31,119,180}

\begin{axis}[
height=\figureheight,
legend cell align={left},
legend style={
  fill opacity=0.8,
  draw opacity=1,
  text opacity=1,
  at={(0.03,0.97)},
  anchor=north west,
  draw=lightgray204
},
tick align=outside,
tick pos=left,
width=\figurewidth,
x grid style={darkgray176},
xlabel={\(\displaystyle \leftarrow\) Size [MB]},
xmajorgrids,
xmin=0, xmax=10.6777473888889,
xtick style={color=black},
y grid style={darkgray176},
ylabel={PSNR \(\displaystyle \rightarrow\)},
ymajorgrids,
ymin=27.5154928525289, ymax=27.8757083786858,
ytick style={color=black}
]
\addplot [semithick, steelblue31119180, mark=*, mark size=3, mark options={solid}]
table {%
4.86701311111111 27.6060299343533
5.85972944444444 27.7556465996636
10.397552 27.8593349456787
};
\addlegendentry{\ours}
\addplot [semithick, darkorange25512714, mark=square*, mark size=3, mark options={solid}]
table {%
4.79364422222222 27.531866285536
6.21614555555556 27.711723751492
9.13027322222222 27.7586309644911
};
\addlegendentry{\ours w/o PE}
\end{axis}

\end{tikzpicture}}\\[-4pt]
  \definecolor{steelblue31119180}{RGB}{31,119,180}
  \definecolor{lightgrey}{RGB}{211,211,211}
  \captionof{figure}{Ablation study on the positional encoding (PE) module. Our proposed PE improves the trade-off between PSNR and model size on \mipnerf.}
  \label{fig:ablation_pe}
  \vspace*{-4em}
\end{wrapfigure}

\paragraph{Ablation studies}
\label{sec:ablations}
The ablation study in \cref{fig:ablation_pe} compares our model with/without positional encoding (PE). The proposed PE module improves the reconstruction quality under the same compression ratio through locality awareness. The ablation study in \cref{table:ablation_compression_modules} compares our model with/without each compression module, showing how much each module contributes to the overall performance.
Additional ablation studies on the impact of model hyperparameters, including recursion depth $R$, feature dimension $n_{f}$, compaction strength $\lambda_o$, and arithmetic coding strength $\lambda_q$, are included in \cref{app:ablation_study}.

\section{Discussion and conclusion}
\label{sec:discussion}
\paragraph{Conclusion} We introduce \ours, a compact 3DGS representation that keeps geometry \emph{explicit} and efficiently packed, and abstracts view-dependent appearance into low-dimensional, arithmetically coded splat features with locality awareness. \ours achieves the highest compression ratio among existing baseline models and maintains competitive reconstruction quality. The compact storage and fast inference also facilitate real-time rendering and transmission of 3D scenes in resource-constrained environments. Extensive experiments on standard 3D reconstruction data sets validate the effectiveness of \ours through quantitative and qualitative experiments. 

\paragraph{Limitations} \ours requires training for each scene, which is time-consuming and limits its scalability to large-scale applications. The positional encoding relies on the octree hierarchy, which is sensitive to splat edits and suboptimal for large-scale scenes.

\paragraph{Future work} The discrete representations of splat features in \ours could serve as 3D information tokens, enabling integration with other learning modalities and language models. The discrete primitive distribution can potentially be improved by depth estimation techniques. The octree hierarchy implies relations among splats, which can improve representation by leveraging prior knowledge of the geometry.

\clearpage  %
\bibliographystyle{unsrtnat}

\clearpage
\setcounter{page}{1}
\appendix

\renewcommand{\thetable}{A\arabic{table}}
\renewcommand{\thefigure}{A\arabic{figure}}

\section{Implementation details}
\label{app:implementation_details}

\subsection{Hardware and software}
\label{app:hardware}
Our experiments are conducted on NVIDIA H200 GPUs with 141~GB of memory. %
The code is implemented in Python 3.11 using the PyTorch framework (version 2.4.1) \cite{paszke2019pytorch} with CUDA~12.4. The Gaussian rasterization module is built on top of the Inria 3DGS repository \cite{kerbl20233dgs}. The arithmetic coding and hash encoding modules are built upon the implementations from HAC \cite{chen2024hac}. The average training time of \ours-base is around 26.58~minutes (35k iterations) per scene.

\subsection{Optimization details}
\paragraph{Learning rates and schedulers}
The model parameters are optimized using the Adam optimizer \cite{kingma2014adam}. We use an exponential decay scheduler on learning rates of abstract features $\vf$, coordinates $\vx$, scaling controller $\vs$, all MLPs ($\MLP_o, \MLP_c, \MLP_r, \MLP_s, \MLP_h, \MLP_{oct}$), and the hash encoder $\mathrm{Hash}(\cdot)$. For each parameter, the learning rate $\eta_i$ at iteration $i$ is defined as
\begin{equation}
\label{eq:lr_schedule}
    \eta_i = \exp((1-t)\log(\eta_0) + t~\log(\eta_{\mathrm{end}})), \; \mathrm{where} \quad t = \mathrm{clamp}(\frac{i}{I}, \min = 0, \max =1).
\end{equation}
Here $I$ is the maximum iteration for the learning rate scheduler, $\eta_0$ and $\eta_{\mathrm{end}}$ are the initial and final learning rates, respectively. We use $I=30,000$, and the details of $\eta_0$ and $\eta_{\mathrm{end}}$ for different model parameters are shown in \cref{table:learning_rate_schedule}.

\begin{table*}[h!]
  \centering\footnotesize
  \caption{\textbf{Learning schedulers} of model parameters. Here $\eta_0$ and $\eta_{\mathrm{end}}$ denote the initial and final learning rates, and the learning rate follows an exponential decay schedule defined in \cref{eq:lr_schedule}.}
  \label{table:learning_rate_schedule}
  \setlength{\tabcolsep}{5pt}
  \resizebox{\textwidth}{!}{
\begin{tabular}{l|cccccccccc}
\toprule
 & $\vx$ & $\vf$ & $\vs$ & $\MLP_o$ & $\MLP_c$ & $\MLP_r$ & $\MLP_s$ & $\MLP_h$ & $\MLP_{oct}$ & $\mathrm{Hash}$ \\
\midrule
$\eta_0$ & 0.0002 & 0.0075 & 0.01 & 0.002 & 0.008 & 0.004 & 0.004 & 0.005 & 0.008 & 0.005 \\
$\eta_{\mathrm{end}}$ & 1$\!\times\! 10^{-5}$ & 0.0075 & 0.002 & 2$\!\times\! 10^{-5}$ & 5$\!\times\! 10^{-5}$ & 0.004 & 0.004 & 1$\!\times\! 10^{-5}$ & 0.008 & 1$\!\times\! 10^{-5}$ \\
\bottomrule
\end{tabular}

  }
\end{table*}
\subsection{Adaptive density control settings}
During the {\em Densification stage} (0.5k--15k iterations), the pruning threshold on opacity is set to 
$\tau_{o} = 0.005$.
The densification threshold on the magnitude of coordinate gradients is defined as 
 $\tau_{g} = 2{\!\times\!}10^{-4}$. During the {\em Compaction stage} (15k--20k iterations), there is no densification module applied, and the pruning threshold on opacity is set to $\tau_{o} = 0.005$.

\begin{figure}[t!]
  \centering\tiny
  \setlength{\figurewidth}{.45\textwidth}
  \setlength{\figureheight}{\figurewidth}
  \hfill
  \begin{minipage}[t]{.5\textwidth}
\begin{tikzpicture}

\definecolor{darkgray176}{RGB}{176,176,176}
\definecolor{darkorange25512714}{RGB}{255,127,14}
\definecolor{forestgreen4416044}{RGB}{44,160,44}
\definecolor{gray}{RGB}{128,128,128}
\definecolor{lightgray204}{RGB}{204,204,204}
\definecolor{steelblue31119180}{RGB}{31,119,180}

\begin{axis}[
axis y line*=right,
height=\figureheight,
tick align=outside,
width=\figurewidth,
x grid style={darkgray176},
xmin=9.26, xmax=16.74,
xtick pos=left,
xtick style={color=black},
y grid style={darkgray176},
ylabel={Splat number ratio (\(\displaystyle \%\))},
ymin=0, ymax=103.843811913113,
ytick pos=right,
ylabel near ticks,
ytick style={color=black},
yticklabel style={anchor=west}
]
\draw[draw=none,fill=gray,fill opacity=0.5] (axis cs:9.6,0) rectangle (axis cs:10.4,16.8604379110568);
\draw[draw=none,fill=gray,fill opacity=0.5] (axis cs:10.6,0) rectangle (axis cs:11.4,36.1764860275442);
\draw[draw=none,fill=gray,fill opacity=0.5] (axis cs:11.6,0) rectangle (axis cs:12.4,61.6683527183754);
\draw[draw=none,fill=gray,fill opacity=0.5] (axis cs:12.6,0) rectangle (axis cs:13.4,82.3385707427946);
\draw[draw=none,fill=gray,fill opacity=0.5] (axis cs:13.6,0) rectangle (axis cs:14.4,92.8173484625893);
\draw[draw=none,fill=gray,fill opacity=0.5] (axis cs:14.6,0) rectangle (axis cs:15.4,97.1388110259468);
\draw[draw=none,fill=gray,fill opacity=0.5] (axis cs:15.6,0) rectangle (axis cs:16.4,98.8988684886793);
\end{axis}

\begin{axis}[
height=\figureheight,
legend cell align={left},
legend style={
  fill opacity=0.8,
  draw opacity=1,
  text opacity=1,
  at={(0.97,0.03)},
  anchor=south east,
  draw=lightgray204
},
tick align=outside,
tick pos=left,
width=\figurewidth,
x grid style={darkgray176},
xlabel={Recursion depth, $r$},
xmin=9.26, xmax=16.74,
xtick style={color=black},
y grid style={darkgray176},
ylabel={PSNR (dB)},
ytick={14,18,22,26},
ymin=13.1662574887276, ymax=28.0489352265994,
ytick style={color=black},
ylabel near ticks
]
\addplot [semithick, steelblue31119180, opacity=1, dashed]
table {%
10 27.3724498748779
11 27.3724498748779
12 27.3724498748779
13 27.3724498748779
14 27.3724498748779
15 27.3724498748779
16 27.3724498748779
};
\addlegendentry{3DGS-30k}
\addplot [semithick, darkorange25512714, opacity=0.8, mark=square*, mark size=3, mark options={solid}]
table {%
10 13.842742840449
11 16.9485883315404
12 20.950058221817
13 24.6404952208201
14 26.5374493598938
15 27.166176478068
16 27.3317444324493
};
\addlegendentry{Quantized-30k}
\addplot [semithick, forestgreen4416044, opacity=0.8, mark=triangle*, mark size=3, mark options={solid}]
table {%
10 23.3284567991892
11 25.13036998113
12 26.5397605101267
13 27.1276112397512
14 27.2863083680471
15 27.3395247459412
16 27.3542562325795
};
\addlegendentry{Quantized-45k}
\end{axis}

\end{tikzpicture}\\[1em]
  \end{minipage}
  \hfill
  \begin{minipage}[t]{.4\textwidth}
  \begin{tikzpicture}[inner sep=0]
    \foreach \recursion/\psnr/\i/\j [count=\k from 0] in {10/13.8427/0/0,12/20.9500/1/0,14/26.5374/0/1,16/27.3317/1/1} {        
      \node (k\k) at (\i*0.5*\textwidth,-\j*0.55*\textwidth) {\includegraphics[width=0.48\textwidth,trim=200 0 200 0,clip]{\figpath/effect_of_recursion/recursion\recursion_render_quantize30k_garden}};
      \node[anchor=north,font=\scriptsize\strut] at (k\k.south) {$r=\recursion$};
\fill[inner sep=0,fill=white,fill opacity=0.7] (k\k.north west) -- ++(2.7em,0) [rounded corners=1pt]-- ++(0,-.8em) [sharp corners]-- ++(-2.7em,0) -- cycle;
\node[anchor=north west, color=black, font=\tiny,inner sep=1pt] at (k\k.north west) {\pgfmathprintnumber[fixed,precision=3,zerofill]{\psnr}};
    }
  \end{tikzpicture}
  \end{minipage}
  \hfill
  \vspace*{-1.5em}
  \caption{\textbf{Quantization recursion depth $r$ vs.\ PSNR.} We experiment with quantizing the coordinates of a trained 3DGS-30k model to be the `Quantized-30k' models. These models are fine-tuned with quantized coordinates fixed for an additional 15k iterations to be `Quantized-45k'. Left: Quantitative metrics (PSNR) and splat number ratio after quantization vs.\ quantization recursion on the {\it Garden} scene. Right: Qualitative results of Quantized-30k. Higher recursions yield better quality but keep more splats.}
  \label{fig:effect_of_recursion_quantitative}
\end{figure}

\subsection{Model hyperparameters}
\label{app:model_hyperparameters}
The feature dimension is fixed to $n_f=8$, and the number of recursions of occupancy-octree for coordinate compression is set to $R=16$. The parameter selection is based on the ablation experiments in \cref{app:ablation_study}.

\paragraph{MLP architectures}
The MLP architecture metadata, including the input dimension, output dimension, hidden dimension, number of layers, and activation types of the hidden layer and output layer, is summarized in \cref{table:mlp_architecture}. The input dimensions of $\MLP_o, \MLP_c, \MLP_r, \MLP_s$ are $n_f + n_{oct} + 4$, which is defined by the concatenation of the abstract feature $\vf$ (dimension $n_f=8$), the output dimension $n_{oct}$ of the octree encoder $\MLP_{oct}$, the 3D viewing direction, and the viewing distance. Here $n_{oct}$ is fixed to be 8.

\paragraph{Hash encoder}
The hash encoder $\mathrm{Hash}(\cdot)$ contains both 2D and 3D hash tables, with size of $2^{15}$ and $2^{13}$, respectively. There are 4-dimensional features per level. The resolution lists of 2D and 3D hash tables are defined as $(130, 258, 514, 1026)$ and $(18, 24, 33, 44, 59, 80, 108, 148, 201, 275, 376, 514)$.

\begin{table*}[h!]
  \centering\footnotesize
  \caption{\textbf{Details of the MLP architectures} in \ours. Here $n_f$ denotes the feature dimension.}
  \label{table:mlp_architecture}
  \setlength{\tabcolsep}{5pt}
    \resizebox{\textwidth}{!}{
\begin{tabular}{l|ccccc}
\toprule
 & Input Dim & Layers & Hidden Dim & Output Dim & Activation(Hidden/Output) \\
\midrule
$\MLP_o$ & $n_f + n_{oct} + 4$ & 3 & 128 & 1 & ReLU/Tanh \\
$\MLP_c$ & $n_f + n_{oct}+ 4$ & 3 & 128 & 3 & ReLU/Sigmoid \\
$\MLP_r$ & $n_f + n_{oct}  + 4$ & 3 & 128 & 4 & ReLU/None \\
$\MLP_s$ & $n_f + n_{oct}  + 4$ & 3 & 128 & 3 & ReLU/None \\
$\MLP_h$ & 96 & 3 & 128 & $8+2n_f$ & ReLU/None \\
$\MLP_{oct}$ & $3r$ & 3 & 64 & $n_{oct}$ & ReLU/None \\
\bottomrule
\end{tabular}
    }
\end{table*}

\paragraph{Model hyperparameters}
For the default settings (\ours-base), the regularization hyperparameters are set to $\lambda_o = 3\!\times \!10^{-7}$, $\lambda_q = 5\!\times \!10^{-4}$ and $\lambda_s = 0.2$, and the recursion depth $r=16$, where $\lambda_o, \lambda_q$ and $\lambda_s$ denote the weights for opacity regularization, quantization regularization, and SSIM-loss regularization, respectively. \looseness-1

\section{Additional experiments}
\label{app:experiments}
In this section, we provide ablation studies on different model hyperparameters and optimization hyperparameters in \cref{app:ablation_study}, additional quantitative results in \cref{app:additional_quantitative_results}, additional qualitative results in \cref{app:additional_qualitative_results}, occupancy-octree analysis in \cref{app:octree_analysis}, encoding and decoding time on each model component per scene in \cref{app:encoding_decoding_details}, and the interpretation of learned abstract features in \cref{app:feat-visualization}.

\subsection{Ablation studies}
\label{app:ablation_study}
We conduct ablation studies on four different hyperparameters in \ours, including recursion level $R$, feature dimension $n_f$, the regularization weights on opacity and quantization $\lambda_o, \lambda_q$. 

\paragraph{Recursion level \boldsymbol{$R$}} 
The value of the total recursion depth $R$ controls the compression ratio of splat coordinates. The ablation experiment on recursion level $R$ is presented in \cref{table:ablation_recursion_levels,fig:ablation_recursion_levels}, where $R \in \{14, 15, 16, 17, 18\}$ in the experiments. Larger $R$ requires linearly more storage memory of coordinates, and leads to better reconstruction quality. Recursion level $R=16$ provides a good trade-off between compression ratio and reconstruction quality. 

\paragraph{Feature dimension \boldsymbol{$n_f$}} 
The value of $n_f$ controls the dimensionality of abstract features.  The ablation experiment on feature dimension $n_f$ is presented in \cref{table:ablation_feature_dimension,fig:ablation_feature_dimension}, where $n_f \in \{ 4, 6, 8, 16, 24\}$ in the experiments. Larger $n_f$ requires more storage space for features, while improving the reconstruction quality. We choose feature dimension $n_f=8$ for compact storage and competitive reconstruction quality, as higher dimensionality yields marginal reconstruction gains while substantially increasing storage costs.

\paragraph{Opacity regularization \boldsymbol{$\lambda_o$}} 
The value of $\lambda_o$ controls the strength of opacity regularization. The ablation experiment on opacity regularization strength $\lambda_o$ is presented in \cref{table:ablation_opacity_regularization,fig:ablation_opacity_regularization}. The value of $\lambda_o$ is set to $\{1\!\times\! 10^{-7}, 2\!\times\! 10^{-7}, 3\!\times\! 10^{-7}, 4\!\times\! 10^{-7}, 5\!\times\! 10^{-7}\}$ in the experiments. The larger $\lambda_o$ encourages the model to use fewer splats to represent the scene, resulting in worse reconstruction quality but smaller storage size. And $\lambda_o \in [3\!\times\! 10^{-7}, 4\!\times\! 10^{-7}]$ maintains proper reconstruction quality. Thus, we select our model variants in this range.

\paragraph{Quantization regularization \boldsymbol{$\lambda_q$}} 
The value of $\lambda_q$ controls the strength of quantization regularization, which encourages the model to use fewer bits to represent the abstract features $\vf$ and scaling controller $\vs$. The ablation experiment on quantization regularization strength $\lambda_q$ is presented in \cref{table:ablation_quantization_regularization,fig:ablation_quantization_regularization}, where $\lambda_q \in \{2\!\times \!10^{-4}, 5\!\times \!10^{-4}, 1\!\times \!10^{-3}, 2\!\times \!10^{-3}, 3\!\times \!10^{-3}\}$ in the experiments. Larger $\lambda_q$ encourages the model to use fewer bits for feature representation $\vf$ and scaling controller $\vs$, which sacrifices reconstruction quality and reduces the model storage size slowly when $\lambda_q \geq 5 \!\times\! 10^{-4}$. Therefore, we choose $\lambda_q = 5\!\times \!10^{-4}$ for our default model \ours-base.

\begin{table*}[t!]
  \footnotesize\centering
  \caption{\textbf{Ablation study on recursion level \boldsymbol{$R$} of occupancy-octree encoding}. The results show that $R=16$ provides sufficient reconstruction quality.}
  \label{table:ablation_recursion_levels}
  \setlength{\tabcolsep}{5pt}
  \newcommand{\rotate}[2]{\parbox[c]{2mm}{\multirow{#1}{*}{\tikz[baseline]\node[anchor=base,rotate=90,font=\tiny,align=center]{#2};}}}
  \resizebox{\textwidth}{!}{%
\begin{tabular}{l|cccc|cccc|cccc}
    \toprule
    & \multicolumn{4}{c}{\mipnerf} & \multicolumn{4}{c}{\tandt} & \multicolumn{4}{c}{\deepblending}
    \\ 
    Recursion Level& PSNR$\uparrow$ & SSIM$\uparrow$ & LPIPS$\downarrow$ & Size$\downarrow$ &PSNR$\uparrow$ & SSIM$\uparrow$ & LPIPS$\downarrow$ & Size$\downarrow$ & PSNR$\uparrow$ & SSIM$\uparrow$ & LPIPS$\downarrow$ & Size$\downarrow$
    \\\midrule
$R=14$ & 27.36 & 0.787 & 0.265 & 5.132 & 23.58 & 0.808 & 0.244 & 3.769 & 29.69 & 0.901 & 0.272 & 2.789 \\
$R=15$ & 27.60 & 0.801 & 0.252 & 5.513 & 24.07 & 0.829 & 0.220 & 4.463 & 30.17 & 0.906 & 0.260 & 2.798 \\
$R=16$ & 27.69 & 0.806 & 0.248 & 5.896 & 24.23 & 0.842 & 0.201 & 4.8 & 30.05 & 0.905 & 0.263 & 2.747 \\
$R=17$ & 27.68 & 0.807 & 0.248 & 6.084 & 24.32 & 0.846 & 0.199 & 4.813 & 30.05 & 0.905 & 0.262 & 3.114 \\
$R=18$ & 27.70 & 0.808 & 0.248 & 6.207 & 24.28 & 0.847 & 0.195 & 4.958 & 30.07 & 0.904 & 0.264 & 2.851 \\
\bottomrule
    \end{tabular} %
}
\end{table*}

\begin{figure*}[t!]
\centering\small
  \pgfplotsset{
    y tick label style={rotate=90},
    semithick/.append style={line width=3pt,opacity=.8},
    legend style={font=\large},
    ylabel style={font=\large},
    xlabel style={font=\large},
    grid style={densely dotted,lightgray!50},
    legend style={draw=none,inner xsep=2pt, inner ysep=0.5pt, nodes={inner sep=1.5pt, text depth=0.1em},fill=white,fill opacity=0.8}
    }
    \foreach \metric [count=\i] in {PSNR, SSIM, LPIPS, Size} {
        \raggedleft 
        \scalebox{.4}{\input{\figpath/app_ablation/recursion/recursion_\metric.tex}}
  }
  \vspace*{-.25em}
  \definecolor{steelblue31119180}{RGB}{31,119,180}
  \definecolor{lightgrey}{RGB}{211,211,211}
  \caption{\textbf{Ablation study on recursion level \boldsymbol{$R$} of occupancy-octree encoding}. In each subfigure, the $x$-axis represents different recursion levels $R$, and the $y$-axis represents the corresponding metric values: PSNR, SSIM, LPIPS, and Size (in MBs).}
  \label{fig:ablation_recursion_levels}
  \vspace*{-4pt}
\end{figure*}

\begin{table*}[t!]
  \footnotesize\centering
  \caption{\textbf{Ablation study on feature dimension \boldsymbol{$n_f$}.} $n_f\!=\!8$ provides a good trade-off between size and quality.}
  \label{table:ablation_feature_dimension}
  \setlength{\tabcolsep}{5pt}
  \newcommand{\rotate}[2]{\parbox[c]{2mm}{\multirow{#1}{*}{\tikz[baseline]\node[anchor=base,rotate=90,font=\tiny,align=center]{#2};}}}
  \resizebox{\textwidth}{!}{%
\begin{tabular}{l|cccc|cccc|cccc}
    \toprule
    & \multicolumn{4}{c}{\mipnerf} & \multicolumn{4}{c}{\tandt} & \multicolumn{4}{c}{\deepblending}
    \\ 
    Feature Dimension& PSNR$\uparrow$ & SSIM$\uparrow$ & LPIPS$\downarrow$ & Size$\downarrow$ &PSNR$\uparrow$ & SSIM$\uparrow$ & LPIPS$\downarrow$ & Size$\downarrow$ & PSNR$\uparrow$ & SSIM$\uparrow$ & LPIPS$\downarrow$ & Size$\downarrow$
    \\\midrule
$n_f=4$ & 27.37 & 0.799 & 0.255 & 5.038 & 23.91 & 0.835 & 0.207 & 4.054 & 30.07 & 0.905 & 0.263 & 2.64 \\
$n_f=6$ & 27.61 & 0.803 & 0.252 & 5.036 & 24.06 & 0.835 & 0.212 & 4.2 & 30.03 & 0.903 & 0.267 & 2.316 \\
$n_f=8$ & 27.69 & 0.806 & 0.248 & 5.896 & 24.23 & 0.842 & 0.201 & 4.8 & 30.05 & 0.905 & 0.263 & 2.747 \\
$n_f=16$ & 27.80 & 0.809 & 0.245 & 8.398 & 24.30 & 0.845 & 0.200 & 6.663 & 30.11 & 0.905 & 0.262 & 3.89 \\
$n_f=24$ & 27.84 & 0.809 & 0.245 & 10.24 & 24.19 & 0.838 & 0.209 & 6.354 & 30.10 & 0.905 & 0.264 & 3.775 \\
\bottomrule
    \end{tabular} %
}
\end{table*}
\begin{figure*}[!]
\centering\small
  \pgfplotsset{
    y tick label style={rotate=90},
    semithick/.append style={line width=3pt,opacity=.8},
    legend style={font=\large},
    ylabel style={font=\large},
    xlabel style={font=\large},
    grid style={densely dotted,lightgray!50},
    legend style={draw=none,inner xsep=2pt, inner ysep=0.5pt, nodes={inner sep=1.5pt, text depth=0.1em},fill=white,fill opacity=0.8}
    }
    \foreach \metric [count=\i] in {PSNR, SSIM, LPIPS, Size} {
        \raggedleft 
        \scalebox{.4}{\input{\figpath/app_ablation/featdim/featdim_\metric.tex}}
  }
  \vspace*{-.25em}
  \definecolor{steelblue31119180}{RGB}{31,119,180}
  \definecolor{lightgrey}{RGB}{211,211,211}
  \caption{\textbf{Ablation study on feature dimension \boldsymbol{$n_f$}.} In each subfigure, the $x$-axis represents different feature dimensions $n_f$, and the $y$-axis represents the corresponding metric values: PSNR, SSIM, LPIPS, and Size (in MBs).}
  \label{fig:ablation_feature_dimension}
\end{figure*}

\begin{table*}[t!]
  \footnotesize\centering
  \caption{\textbf{Ablation study on opacity regularization strength \boldsymbol{$\lambda_o$}.} We choose $
  \lambda_o = 3\!\times\! 10^{-7}$ for the trade-off on reconstruction quality and model size. }
  \label{table:ablation_opacity_regularization}
  \setlength{\tabcolsep}{5pt}
  \newcommand{\rotate}[2]{\parbox[c]{2mm}{\multirow{#1}{*}{\tikz[baseline]\node[anchor=base,rotate=90,font=\tiny,align=center]{#2};}}}
  \resizebox{\textwidth}{!}{%
\begin{tabular}{l|cccc|cccc|cccc}
    \toprule
    & \multicolumn{4}{c}{\mipnerf} & \multicolumn{4}{c}{\tandt} & \multicolumn{4}{c}{\deepblending}
    \\ 
    Opacity Regularization& PSNR$\uparrow$ & SSIM$\uparrow$ & LPIPS$\downarrow$ & Size$\downarrow$ &PSNR$\uparrow$ & SSIM$\uparrow$ & LPIPS$\downarrow$ & Size$\downarrow$ & PSNR$\uparrow$ & SSIM$\uparrow$ & LPIPS$\downarrow$ & Size$\downarrow$
    \\\midrule
$\lambda=1e-7$ & 27.85 & 0.812 & 0.231 & 10.31 & 24.30 & 0.846 & 0.189 & 7.11 & 30.16 & 0.906 & 0.253 & 4.678 \\
$\lambda=2e-7$ & 27.78 & 0.810 & 0.239 & 7.563 & 24.16 & 0.842 & 0.196 & 5.596 & 30.16 & 0.906 & 0.258 & 3.396 \\
$\lambda=3e-7$ & 27.69 & 0.806 & 0.248 & 5.896 & 24.23 & 0.842 & 0.201 & 4.8 & 30.05 & 0.905 & 0.263 & 2.747 \\
$\lambda=4e-7$ & 27.59 & 0.801 & 0.258 & 4.825 & 24.17 & 0.840 & 0.204 & 4.19 & 30.07 & 0.904 & 0.265 & 2.458 \\
$\lambda=5e-7$ & 27.44 & 0.794 & 0.270 & 4.033 & 24.07 & 0.837 & 0.209 & 3.951 & 30.00 & 0.903 & 0.268 & 2.23 \\
\bottomrule
    \end{tabular} %
}
\end{table*}
\begin{figure*}[!]
\centering\small
  \pgfplotsset{
    y tick label style={rotate=90},
    semithick/.append style={line width=3pt,opacity=.8},
    legend style={font=\large},
    ylabel style={font=\large},
    xlabel style={font=\large},
    grid style={densely dotted,lightgray!50},
    legend style={draw=none,inner xsep=2pt, inner ysep=0.5pt, nodes={inner sep=1.5pt, text depth=0.1em},fill=white,fill opacity=0.8}
    }
    \foreach \metric [count=\i] in {PSNR, SSIM, LPIPS, Size} {
        \raggedleft 
        \scalebox{.39}{\input{\figpath/app_ablation/rego/rego_\metric.tex}}
  }
  \vspace*{-.25em}
  \definecolor{steelblue31119180}{RGB}{31,119,180}
  \definecolor{lightgrey}{RGB}{211,211,211}
  \caption{\textbf{Ablation study on opacity regularization strength \boldsymbol{$\lambda_o$}.} In each subfigure, the $x$-axis represents different opacity regularization strength $\lambda_o$, and the $y$-axis represents the corresponding metric values: PSNR, SSIM, LPIPS, and Size (in MBs).}
  
  \label{fig:ablation_opacity_regularization}
\end{figure*}

\begin{table*}[t!]
  \footnotesize\centering
  \caption{\textbf{Ablation study on quantization regularization strength \boldsymbol{$\lambda_q$}.} We select $\lambda_q = 5\!\times\! 10^{-4}$ for the best marginal effect on overall metrics.}
  \label{table:ablation_quantization_regularization}
  \setlength{\tabcolsep}{5pt}
  \newcommand{\rotate}[2]{\parbox[c]{2mm}{\multirow{#1}{*}{\tikz[baseline]\node[anchor=base,rotate=90,font=\tiny,align=center]{#2};}}}
  \resizebox{\textwidth}{!}{%
\begin{tabular}{l|cccc|cccc|cccc}
    \toprule
    & \multicolumn{4}{c}{\mipnerf} & \multicolumn{4}{c}{\tandt} & \multicolumn{4}{c}{\deepblending}
    \\ 
    Quantization Regularization& PSNR$\uparrow$ & SSIM$\uparrow$ & LPIPS$\downarrow$ & Size$\downarrow$ &PSNR$\uparrow$ & SSIM$\uparrow$ & LPIPS$\downarrow$ & Size$\downarrow$ & PSNR$\uparrow$ & SSIM$\uparrow$ & LPIPS$\downarrow$ & Size$\downarrow$
    \\\midrule
$\lambda_q=2e-4$ & 27.73 & 0.807 & 0.248 & 6.516 & 23.67 & 0.821 & 0.233 & 4.666 & 30.10 & 0.905 & 0.262 & 3.034 \\
$\lambda_q=5e-4$ & 27.69 & 0.806 & 0.248 & 5.896 & 24.23 & 0.842 & 0.201 & 4.8 & 30.05 & 0.905 & 0.263 & 2.747 \\
$\lambda_q=1e-3$ & 27.67 & 0.806 & 0.249 & 5.454 & 24.22 & 0.841 & 0.201 & 4.467 & 30.06 & 0.904 & 0.262 & 2.651 \\
$\lambda_q=2e-3$ & 27.53 & 0.804 & 0.251 & 5.065 & 24.08 & 0.839 & 0.203 & 4.089 & 29.97 & 0.903 & 0.265 & 2.407 \\
$\lambda_q=3e-3$ & 27.42 & 0.802 & 0.253 & 4.797 & 24.08 & 0.838 & 0.205 & 3.93 & 29.52 & 0.898 & 0.281 & 2.664 \\
\bottomrule
    \end{tabular} %
}
\end{table*}

\begin{figure*}[t!]
\centering\small
  \pgfplotsset{
    y tick label style={rotate=90},
    semithick/.append style={line width=3pt,opacity=.8},
    legend style={font=\large},
    ylabel style={font=\large},
    xlabel style={font=\large},
    grid style={densely dotted,lightgray!50},
    legend style={draw=none,inner xsep=2pt, inner ysep=0.5pt, nodes={inner sep=1.5pt, text depth=0.1em},fill=white,fill opacity=0.8}
    }
    \foreach \metric [count=\i] in {PSNR, SSIM, LPIPS, Size} {
        \raggedleft 
        \scalebox{.39}{\input{\figpath/app_ablation/regbit/regbit_\metric.tex}}
  }
  \vspace*{-.25em}
  \definecolor{steelblue31119180}{RGB}{31,119,180}
  \definecolor{lightgrey}{RGB}{211,211,211}
  \caption{\textbf{Ablation study on quantization regularization strength \boldsymbol{$\lambda_q$}.} In each subfigure, the $x$-axis represents different quantization regularization strength $\lambda_q$, and the $y$-axis represents the corresponding metric values: PSNR, SSIM, LPIPS, and Size (in MBs).}
  \label{fig:ablation_quantization_regularization}
\end{figure*}

\subsection{Additional quantitative results}
\label{app:additional_quantitative_results}
The overall quantitative comparison among all baselines on three benchmark data sets is presented as a scatter plot in \cref{fig:quantitative_benchmark_scatter}. The results demonstrate that \ours achieves competitive reconstruction quality with the lowest memory footprints compared to other baselines.

\begin{figure*}[t!]
\centering\scriptsize
  \setlength{\figurewidth}{.5\textwidth}
  \setlength{\figureheight}{.5\textwidth}
  \pgfplotsset{
    y tick label style={rotate=90},
    semithick/.append style={line width=3pt,opacity=.8},
    legend style={font=\footnotesize},
    ylabel style={font=\large},
    xlabel style={font=\large},
    grid style={densely dotted,lightgray!50},
    legend style={draw=none,inner xsep=2pt, inner ysep=0.5pt, nodes={inner sep=1.5pt, text depth=0.1em},fill=white,fill opacity=0.8}
    }
  \foreach \dataset/\nicename [count=\j] in {MipNeRF360/{\mipnerf}, TanksAndTemples/{\tandt}, DeepBlending/{\deepblending}} {
    \begin{subfigure}[b]{0.025\textwidth}
      \centering 
      \tikz\node[rotate=90,text width=.8\figureheight,align=center,font=\bf\strut,scale=.8]{\nicename};
    \end{subfigure}%
    \foreach \metric [count=\i] in {PSNR, SSIM, LPIPS} {
      \begin{subfigure}[b]{0.31\textwidth}
        \raggedleft 
        \scalebox{.6}{\input{\figpath/quantitative_scatter_plot/\dataset_\metric.tex}}
      \end{subfigure}
  }}
  \vspace*{-.25em}
  \definecolor{steelblue31119180}{RGB}{31,119,180}
  \definecolor{lightgrey}{RGB}{211,211,211}
  \caption{\textbf{Comparison of reconstruction quality across baselines.} In each subfigure, the $x$-axis represents the model size (in MBs), and the $y$-axis represents the corresponding metric values: PSNR (higher is better), SSIM (higher is better), LPIPS (lower is better). Different rows correspond to different benchmark data sets: \mipnerf, \tandt, and \deepblending. Each line connects the results of a specific model over different hyperparameters (variants).
  }
  \label{fig:quantitative_benchmark_scatter}
\end{figure*}

\subsection{Additional qualitative results}
\label{app:additional_qualitative_results}
Additional qualitative comparison among ground truth, 3DGS-30K, \ours-base, and \ours-small is presented in \cref{fig:qualitative_all}. The results show that \ours preserves scene details with fewer artefacts than 3DGS-30K, even with smaller model sizes. \ours-small loses some fine details compared to \ours-base, but still maintains competitive visual quality.
\begin{figure*}[!t]
  \centering
  \hspace*{0.03\textwidth}
  \foreach \method [count=\i] in {{Ground truth},{3DGS-30K},{\ours-base},{\ours-small}}{
    \begin{subfigure}[c]{0.22\textwidth}
      \centering\strut\bf\method
    \end{subfigure}
  }
  \foreach \scene/\scenename [count=\j] in {garden/Garden, room/Room, bicycle/Bicycle, kitchen/Kitchen, bonsai/Bonsai, train/Train, truck/Truck, counter/Counter} {
  \begin{tabular}{c}
    \begin{subfigure}[c]{0.03\textwidth}
      \tikz\node[rotate=90,font=\bf\strut]{\itshape \scenename};
    \end{subfigure}
    \foreach \method/\path [count=\i] in {gt/gt,3dgs/3dgs,\ours-base/smolgs-base,\ours-small/smolgs-small}{
      \begin{subfigure}[c]{0.22\textwidth} %
        \centering
        \includegraphics[width=\textwidth]{\figpath/app_qualitative/\path_\scene_00001}
      \end{subfigure}
      
      \vspace*{-1pt}
    }
  \end{tabular}\\[.5em]
  }
  \caption{\textbf{Comparison of qualitative results} among the ground truth, 3DGS-30K, \ours-base, and \ours-small.}
  \label{fig:qualitative_all}
\end{figure*}

\subsection{Occupancy-octree analysis}
\label{app:octree_analysis}
The occupancy-octree learned in \cref{sec:coordinate-compression} generates a set of byte-codes during the encoding process. The statistical characteristics of the byte-codes are analyzed in \cref{fig:histogram_octree}. The results show that most divisions generate fewer non-empty sub-boxes, indicating that the byte-codes are highly sparse and thus compressible by entropy coding. 

\begin{figure*}[t!]
\centering\small
  \setlength{\figurewidth}{.5\textwidth}
  \setlength{\figureheight}{.5\textwidth}
  \pgfplotsset{
    y tick label style={rotate=90},
    semithick/.append style={line width=3pt,opacity=.8},
    legend style={font=\footnotesize},
    ylabel style={font=\large},
    xlabel style={font=\large},
    grid style={densely dotted,lightgray!50},
    legend style={draw=none,inner xsep=2pt, inner ysep=0.5pt, nodes={inner sep=1.5pt, text depth=0.1em},fill=white,fill opacity=0.8}
    }
    \hspace*{0.03\textwidth}
   \foreach \scene/\scenename [count=\i] in {bonsai/Bonsai, room/Room, garden/Garden, drjohnson/DrJohnson} {
    \begin{subfigure}[c]{0.22\textwidth}
      \centering\it\strut\scenename
    \end{subfigure}
  }
  \foreach \path/\nicename [count=\j] in {non_empty_subboxes/1,value_frequency/2} {
    \foreach \scene/\scenename [count=\i] in {bonsai/Bonsai, room/Room, garden/Garden, drjohnson/DrJohnson} {
      \begin{subfigure}[b]{0.23\textwidth}
        \raggedleft 
        \scalebox{.46}{\input{\figpath/histogram_octree/\scene_\path.tex}}
      \end{subfigure}
  }}
  \vspace*{-.25em}
  \definecolor{steelblue31119180}{RGB}{31,119,180}
  \definecolor{lightgrey}{RGB}{211,211,211}
  \caption{Statistical analysis of the occupancy-octree byte-codes on four example scenes: {\it Bonsai, Room, Garden, DrJohnson}. The occupancy-octree generates a series of byte-codes during the encoding process. Each byte-code denotes the existence of a splat in the sub-boxes of the corresponding division. The first row shows the histogram of the number of non-empty sub-boxes per division, and the second row shows the histogram of byte-codes (sorted by frequency). Each column corresponds to a specific scene. The histograms indicate the strong sparsity of the occupancy-octree byte-codes shared by different scenes.}
  \label{fig:histogram_octree}
\end{figure*}

\subsection{Encoding and decoding details}
\label{app:encoding_decoding_details}
Per-scene encoding time, decoding time, and reconstruction quality for both \ours-base and \ours-small are summarized in \cref{table:app_encoding_time_per_scene_base,table:app_encoding_time_per_scene_small,table:app_decoding_time_per_scene_base,table:app_decoding_time_per_scene_small,table:app_reconstruction_quality_per_scene_base,table:app_reconstruction_quality_per_scene_small}.

\begin{table}[h!]
  \centering\footnotesize
  \caption{Encoding time (seconds) of each component of \ours-base per scene.}
  \label{table:app_encoding_time_per_scene_base}
  \setlength{\tabcolsep}{5pt}
  \newcommand{\rotate}[2]{\multirow{#1}{*}{\tikz[baseline]\node[anchor=base,rotate=0,align=center]{#2};}}
  \resizebox{0.8\textwidth}{!}{

    \begin{tabular}{ll|ccccc}
\toprule
 \multicolumn{2}{c}{{\bf Encoding time} (s)} & Total & ${\vx}$ & ${\vf}$ & ${\vs}$ & MLPs \\
\midrule
\rotate{9}{\mipnerf}&{\it Garden} & 3.654 & 0.262 & 2.158 & 1.149 & 0.026 \\
&{\it Bicycle} & 1.845 & 0.152 & 1.044 & 0.595 & 0.022 \\
&{\it Stump} & 2.297 & 0.156 & 1.314 & 0.761 & 0.022 \\
&{\it Bonsai} & 1.401 & 0.162 & 0.755 & 0.432 & 0.020 \\
&{\it Counter} & 1.442 & 0.128 & 0.815 & 0.454 & 0.017 \\
&{\it Kitchen} & 1.473 & 0.107 & 0.839 & 0.479 & 0.014 \\
&{\it Room} & 0.961 & 0.102 & 0.495 & 0.320 & 0.020 \\
&{\it Treehill} & 1.277 & 0.101 & 0.704 & 0.420 & 0.022 \\
&{\it Flowers} & 2.988 & 0.223 & 1.704 & 0.970 & 0.024 \\
\midrule
\rotate{2}{\deepblending}&{\it Drjohnson} & 1.315 & 0.120 & 0.764 & 0.354 & 0.048 \\
&{\it Playroom} & 0.996 & 0.093 & 0.587 & 0.262 & 0.031 \\
\midrule
\rotate{2}{\tandt}&{\it Train} & 1.417 & 0.089 & 0.822 & 0.465 & 0.015 \\
&{\it Truck} & 1.794 & 0.142 & 0.977 & 0.609 & 0.023 \\
\bottomrule
\end{tabular}   }
\end{table}
\begin{table}[h!]
  \centering\footnotesize
  \caption{Encoding time (seconds) of each component of \ours-small per scene.}
  \label{table:app_encoding_time_per_scene_small}
  \setlength{\tabcolsep}{5pt}
  \newcommand{\rotate}[2]{\multirow{#1}{*}{\tikz[baseline]\node[anchor=base,rotate=0,align=center]{#2};}}
  \resizebox{0.8\textwidth}{!}{

    \begin{tabular}{ll|ccccc}
\toprule
 \multicolumn{2}{c}{{\bf Encoding time} (s)} & Total & ${\vx}$ & ${\vf}$ & ${\vs}$ & MLPs \\
\midrule
\rotate{9}{\mipnerf}&{\it Garden} & 2.938 & 0.219 & 1.690 & 0.943 & 0.024 \\
&{\it Bicycle} & 1.558 & 0.125 & 0.869 & 0.510 & 0.021 \\
&{\it Stump} & 1.687 & 0.116 & 0.958 & 0.558 & 0.019 \\
&{\it Bonsai} & 1.218 & 0.115 & 0.678 & 0.384 & 0.016 \\
&{\it Counter} & 1.317 & 0.120 & 0.731 & 0.411 & 0.022 \\
&{\it Kitchen} & 1.420 & 0.135 & 0.794 & 0.434 & 0.022 \\
&{\it Room} & 0.831 & 0.088 & 0.419 & 0.277 & 0.023 \\
&{\it Treehill} & 0.856 & 0.077 & 0.481 & 0.266 & 0.012 \\
&{\it Flowers} & 2.411 & 0.199 & 1.424 & 0.728 & 0.017 \\
\midrule
\rotate{2}{\deepblending}&{\it Drjohnson} & 0.883 & 0.100 & 0.451 & 0.295 & 0.018 \\
&{\it Playroom} & 0.706 & 0.095 & 0.334 & 0.230 & 0.023 \\
\midrule
\rotate{2}{\tandt}&{\it Train} & 1.297 & 0.085 & 0.746 & 0.428 & 0.013 \\
&{\it Truck} & 1.510 & 0.124 & 0.836 & 0.500 & 0.018 \\
\bottomrule
\end{tabular}   }
\end{table}

\begin{table}[h!]
  \centering\footnotesize
  \caption{Decoding time (seconds) of each component of \ours-base per scene.}
  \label{table:app_decoding_time_per_scene_base}
  \setlength{\tabcolsep}{5pt}
  \newcommand{\rotate}[2]{\multirow{#1}{*}{\tikz[baseline]\node[anchor=base,rotate=0,align=center]{#2};}}
  \resizebox{0.8\textwidth}{!}{

    \begin{tabular}{ll|ccccc}
\toprule
 \multicolumn{2}{c}{{\bf Decoding time} (s)} & Total & ${\vx}$ & ${\vf}$ & ${\vs}$ & MLPs \\
\midrule
\rotate{9}{\mipnerf}&{\it Garden} & 5.507 & 0.230 & 3.216 & 2.030 & 0.019 \\
&{\it Bicycle} & 2.882 & 0.126 & 1.692 & 1.042 & 0.017 \\
&{\it Stump} & 3.478 & 0.132 & 2.001 & 1.320 & 0.019 \\
&{\it Bonsai} & 2.118 & 0.114 & 1.210 & 0.757 & 0.026 \\
&{\it Counter} & 2.227 & 0.105 & 1.286 & 0.810 & 0.021 \\
&{\it Kitchen} & 2.288 & 0.095 & 1.314 & 0.859 & 0.015 \\
&{\it Room} & 1.439 & 0.095 & 0.758 & 0.552 & 0.028 \\
&{\it Treehill} & 1.955 & 0.079 & 1.117 & 0.741 & 0.015 \\
&{\it Flowers} & 4.635 & 0.195 & 2.736 & 1.676 & 0.019 \\
\midrule
\rotate{2}{\deepblending}&{\it Drjohnson} & 1.657 & 0.152 & 0.785 & 0.608 & 0.108 \\
&{\it Playroom} & 1.441 & 0.132 & 0.557 & 0.468 & 0.281 \\
\midrule
\rotate{2}{\tandt}&{\it Train} & 2.180 & 0.064 & 1.284 & 0.814 & 0.014 \\
&{\it Truck} & 2.752 & 0.115 & 1.572 & 1.041 & 0.018 \\
\bottomrule
\end{tabular}   }
\end{table}

\begin{table}[h!]
  \centering\footnotesize
  \caption{Decoding time (seconds) of each component of \ours-small per scene.}
  \label{table:app_decoding_time_per_scene_small}
  \setlength{\tabcolsep}{5pt}
  \newcommand{\rotate}[2]{\multirow{#1}{*}{\tikz[baseline]\node[anchor=base,rotate=0,align=center]{#2};}}
  \resizebox{0.8\textwidth}{!}{

    \begin{tabular}{ll|ccccc}
\toprule
 \multicolumn{2}{c}{{\bf Decoding time} (s)} & Total & ${\vx}$ & ${\vf}$ & ${\vs}$ & MLPs \\
\midrule
\rotate{9}{\mipnerf}&{\it Garden} & 4.485 & 0.192 & 2.634 & 1.630 & 0.019 \\
&{\it Bicycle} & 2.420 & 0.105 & 1.400 & 0.890 & 0.019 \\
&{\it Stump} & 2.527 & 0.094 & 1.438 & 0.972 & 0.018 \\
&{\it Bonsai} & 1.869 & 0.099 & 1.068 & 0.681 & 0.016 \\
&{\it Counter} & 1.980 & 0.095 & 1.133 & 0.730 & 0.018 \\
&{\it Kitchen} & 2.097 & 0.104 & 1.202 & 0.758 & 0.025 \\
&{\it Room} & 1.237 & 0.072 & 0.660 & 0.484 & 0.017 \\
&{\it Treehill} & 1.257 & 0.052 & 0.717 & 0.469 & 0.014 \\
&{\it Flowers} & 3.624 & 0.151 & 2.185 & 1.265 & 0.017 \\
\midrule
\rotate{2}{\deepblending}&{\it Drjohnson} & 1.285 & 0.081 & 0.654 & 0.530 & 0.017 \\
&{\it Playroom} & 0.972 & 0.068 & 0.480 & 0.403 & 0.018 \\
\midrule
\rotate{2}{\tandt}&{\it Train} & 2.004 & 0.059 & 1.184 & 0.744 & 0.013 \\
&{\it Truck} & 2.343 & 0.097 & 1.368 & 0.854 & 0.017 \\
\bottomrule
\end{tabular}   }
\end{table}
 
\begin{table}[h!]
  \centering\footnotesize
  \caption{Reconstruction quality per scene for \ours-base.}
  \label{table:app_reconstruction_quality_per_scene_base}
  \setlength{\tabcolsep}{5pt}
  \newcommand{\rotate}[2]{\multirow{#1}{*}{\tikz[baseline]\node[anchor=base,rotate=0,align=center]{#2};}}
  \resizebox{0.8\textwidth}{!}{

    \begin{tabular}{ll|cccc}
\toprule
 \multicolumn{2}{c}{{\bf Reconstruction}} & PSNR$\uparrow$ & SSIM$\uparrow$ & LPIPS$\downarrow$ &  Size$\downarrow$ \\
\midrule
\rotate{9}{\mipnerf}&{\it Garden} & 27.327 & 0.844 & 0.157 & 11.283 \\
&{\it Bicycle} & 24.881 & 0.720 & 0.310 & 5.644 \\
&{\it Stump} & 26.658 & 0.767 & 0.271 & 7.015 \\
&{\it Bonsai} & 32.971 & 0.945 & 0.188 & 4.010 \\
&{\it Counter} & 29.707 & 0.914 & 0.191 & 4.498 \\
&{\it Kitchen} & 31.440 & 0.925 & 0.133 & 4.470 \\
&{\it Room} & 32.088 & 0.924 & 0.206 & 2.858 \\
&{\it Treehill} & 23.210 & 0.635 & 0.396 & 3.782 \\
&{\it Flowers} & 21.519 & 0.587 & 0.379 & 9.178 \\
\midrule
\rotate{2}{\deepblending}&{\it Drjohnson} & 29.627 & 0.902 & 0.263 & 3.331 \\
&{\it Playroom} & 30.617 & 0.907 & 0.261 & 2.390 \\
\midrule
\rotate{2}{\tandt}&{\it Train} & 22.608 & 0.806 & 0.241 & 4.198 \\
&{\it Truck} & 25.889 & 0.879 & 0.157 & 5.369 \\
\bottomrule
\end{tabular}   }
\end{table}

\begin{table}[h!]
  \centering\footnotesize
  \caption{Reconstruction quality per scene for \ours-small.}
  \label{table:app_reconstruction_quality_per_scene_small}
  \setlength{\tabcolsep}{5pt}
  \newcommand{\rotate}[2]{\multirow{#1}{*}{\tikz[baseline]\node[anchor=base,rotate=0,align=center]{#2};}}
  \resizebox{0.8\textwidth}{!}{

    \begin{tabular}{ll|cccc}
\toprule
 \multicolumn{2}{c}{{\bf Reconstruction}} & PSNR$\uparrow$ & SSIM$\uparrow$ & LPIPS$\downarrow$ &  Size$\downarrow$ \\
\midrule
\rotate{9}{\mipnerf}&{\it Garden} & 27.158 & 0.836 & 0.171 & 8.976 \\
&{\it Bicycle} & 24.735 & 0.713 & 0.321 & 4.725 \\
&{\it Stump} & 26.640 & 0.763 & 0.284 & 5.147 \\
&{\it Bonsai} & 32.854 & 0.944 & 0.191 & 3.669 \\
&{\it Counter} & 29.602 & 0.913 & 0.194 & 4.098 \\
&{\it Kitchen} & 31.208 & 0.923 & 0.136 & 4.063 \\
&{\it Room} & 31.912 & 0.922 & 0.210 & 2.530 \\
&{\it Treehill} & 22.868 & 0.616 & 0.430 & 2.696 \\
&{\it Flowers} & 21.477 & 0.581 & 0.389 & 7.900 \\
\midrule
\rotate{2}{\deepblending}&{\it Drjohnson} & 29.565 & 0.901 & 0.267 & 2.777 \\
&{\it Playroom} & 30.374 & 0.906 & 0.265 & 2.067 \\
\midrule
\rotate{2}{\tandt}&{\it Train} & 22.593 & 0.805 & 0.245 & 3.827 \\
&{\it Truck} & 25.823 & 0.877 & 0.163 & 4.590 \\
\bottomrule
\end{tabular}   }
\end{table}

\subsection{Interpretation of features}
\label{app:feat-visualization}
The abstract features learned by \ours-base are visualized in \cref{fig:feat}. The features are visualized using a dimensionality reduction technique and transformation. Given feature $\vf$, the color visualized is defined as $\mathrm{RGB}=\mathrm{sigmoid}(\mathrm{PCA}(\vf)) \in [0,1]^{3}$, where $\mathrm{PCA}$ denotes the principal component analysis. We can observe that the learned features capture meaningful optical cues and semantics.

\begin{figure*}[!t]
  \centering
  \hspace*{0.03\textwidth}
  \foreach \method [count=\i] in {{View 1},{View 1 (feat)},{View 2},{View 2 (feat)}}{
    \begin{subfigure}[c]{0.22\textwidth}
      \centering\strut\bf\method
    \end{subfigure}
  }
  \foreach \scene/\scenename [count=\j] in {garden/Garden, room/Room, bicycle/Bicycle, kitchen/Kitchen, bonsai/Bonsai, train/Train, truck/Truck, counter/Counter} {
  \begin{tabular}{c}
    \begin{subfigure}[c]{0.03\textwidth}
      \tikz\node[rotate=90,font=\bf\strut]{\itshape\scenename};
    \end{subfigure}
    \foreach \method [count=\i] in {00000,feat_00000,00001,feat_00001}{
      \begin{subfigure}[c]{0.22\textwidth} %
        \centering
        \includegraphics[width=\textwidth]{\figpath/app_feature_interp/smolgs-base_\scene_\method}
      \end{subfigure}
      \vspace*{-1pt}
    }
  \end{tabular}\\[.5em]
  }
  \caption{\textbf{Visualization of learned abstract features $\vf$.} For each scene, we show two different views and their corresponding rendering image and feature visualizations. The features are visualized by mapping the 8-dimensional features to RGB colors using $\mathrm{RGB}=\mathrm{sigmoid}(\mathrm{PCA}(\vf))$. }
  \label{fig:feat}
\end{figure*}

\end{document}